%% file: wacv2025_arXiv.tex
\documentclass[10pt,twocolumn,letterpaper]{article}

\usepackage[pagenumbers]{cvpr} 

\usepackage{graphicx}
\usepackage{amsmath}
\usepackage{amssymb}
\usepackage{booktabs}

\usepackage[accsupp]{axessibility} 

\usepackage{subfloat} 
\usepackage{multirow}
\usepackage{tablefootnote}
\usepackage{xcolor,colortbl}
\usepackage{wrapfig}

\input{mlpreamble}

\newcommand{\constraint}{\ensuremath{{E_{tree}}}}
\newcommand{\func}{\ensuremath{\mathcal{F}}}

\newcommand{\Proj}{\ensuremath{\mathcal{P}}}
\newcommand{\Rep}{\ensuremath{\mathcal{R}}}
\newcommand{\Loss}{\ensuremath{\mathcal{L}}}

\newcommand{\img}{\ensuremath{I}}
\newcommand{\graph}{\ensuremath{{G}}}
\newcommand{\ijsub}{\ensuremath{(i,j)}}

\newcommand{\softmax}{\ensuremath{\V{\sigma}}}

\newcommand{\edge}{\ensuremath{{E}}}
\newcommand{\vertex}{\ensuremath{{V}}}
\newcommand{\REP}{SFS\xspace}

\newcommand{\real}{\mathbb{R}}
\newcommand{\unconst}[1]{\ensuremath{\hat{#1}}}

\usepackage[pagebackref,breaklinks,colorlinks]{hyperref}

\usepackage[capitalize]{cleveref}
\crefname{section}{Sec.}{Secs.}
\Crefname{section}{Section}{Sections}
\Crefname{table}{Table}{Tables}
\crefname{table}{Tab.}{Tabs.}

\def\confName{WACV}
\def\confYear{2025}

\begin{document}

\title{TreeFormer: Single-view Plant Skeleton Estimation \\via Tree-constrained Graph Generation}

\author{Xinpeng Liu$^1$
\qquad
Hiroaki Santo$^1$
\qquad
Yosuke Toda$^{2,3}$
\qquad
Fumio Okura$^1$\\
$^1$Osaka University \qquad $^2$Phytometrics \qquad $^3$Nagoya University\\
{\tt\small \{liu.xinpeng,santo.hiroaki,okura\}@ist.osaka-u.ac.jp} \qquad {\tt\small yosuke@phytometrics.jp}
}

\twocolumn[{%
\renewcommand\twocolumn[1][]{#1}%

\maketitle

\begin{center}
	\centering
	\includegraphics[width=\linewidth]{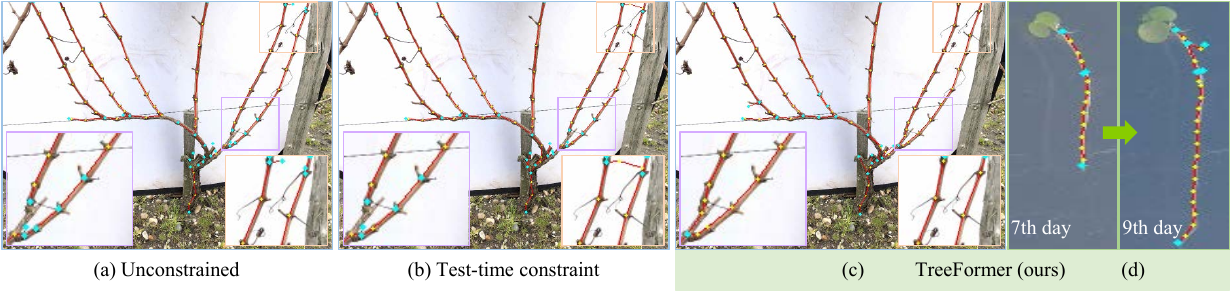}  
     \vspace{-5mm}
    \captionof{figure}{We propose a method for single-image plant skeleton estimation combining learning-based graph generators with traditional graph algorithm (\ie, MST). The red lines show the predicted graph edges. Compared to (a) an \emph{unconstrained} graph generator and (b) a naive tree-graph constraint implementation, (c) our method naturally imposes the constraint during the graph generation models' training. Our method can be directly applied to plant science and agricultural applications, such as (d) time-series reconstruction of botanical roots.}
	\label{fig:intro}
\end{center}
}]

\begin{abstract}
Accurate estimation of plant skeletal structure (\eg, branching structure) from images is essential for smart agriculture and plant science. Unlike human skeletons with fixed topology, plant skeleton estimation presents a unique challenge, \ie, estimating arbitrary tree graphs from images. While recent graph generation methods successfully infer thin structures from images, it is challenging to constrain the output graph strictly to a tree structure. To this problem, we present TreeFormer, a plant skeleton estimator via tree-constrained graph generation. Our approach combines learning-based graph generation with traditional graph algorithms to impose the constraints during the training loop. Specifically, our method \emph{projects} an unconstrained graph onto a minimum spanning tree (MST) during the training loop and incorporates this prior knowledge into the gradient descent optimization by suppressing unwanted feature values. Experiments show that our method accurately estimates target plant skeletal structures for multiple domains: Synthetic tree patterns, real botanical roots, and grapevine branches. Our implementations are available at \url{https://github.com/huntorochi/TreeFormer/}.

\end{abstract}

\section{Introduction}
\label{sec:intro}

Skeletal structures of plants (\eg, branches and roots) are key information for analyzing plant traits in agriculture and plant science.
In particular, single-view estimation of plant skeletons has potential benefits for various downstream tasks, such as high-throughput plant phenotyping~\cite{Guyot, skeleton3dMulitRUE, skeleton3dMaizePoint} and plant organ segmentation~\cite{skeletonSeg, 3dpointSeg}. 
As a similar task, single-view estimation of human poses has been widely studied, \eg, OpenPose~\cite{OpenPose}. However, unlike human skeletons, which have a fixed graph topology, the plant skeleton is not organized because the number of joints and their relationships are unknown, posing a unique problem of estimating an arbitrary tree graph from an image.

The estimation of graph structure from images has been studied to extract thin structures such as road networks in satellite images~\cite{Sat2Graph, SOTAWorkRNGDet, iCurbIL}.
Recent end-to-end models using recurrent neural networks (RNNs)~\cite{polygon-rnn++}, graph neural networks (GNNs)~\cite{StrucLandDetect, PolyBuildSeg, CurveGCN}, or transformers~\cite{GGT, Image2SMILES, SGTR, PolyTransform, Relationformer} show the ability to extract faithful \emph{unconstrained} graph structures from images. 
However, inferring tree-constrained graphs with the existing graph generators becomes a non-trivial problem, where the output graph often violates the required constraints, as shown in \fref{fig:intro}(a).
One reason for this difficulty is that tree graph generation, which requires finding a set of graph edges that satisfy the constraint defined on the entire graph, naturally falls into combinatorial optimization. 
A simple way to impose the constraints on the graph generation is to convert the inferred unconstrained graphs to the closest graph that satisfies the given constraint using traditional graph algorithms such as Dijkstra's shortest path or minimum spanning tree (MST) algorithms.
Such post-processing can work; however, because the graph generators are trained without any constraints, the output may be unrealistic, as shown in \fref{fig:intro}(b).

For tree graph generation from single images, we propose a simple yet effective way to integrate state-of-the-art learning-based graph generation methods, which achieve high-quality image-based graph estimation, and traditional graph algorithms, which compute strictly constrained tree graphs. 
Specifically, we propose to \emph{project} an unconstrained graph into a tree graph by a non-differentiable MST algorithm during each training loop. 
Our selective feature suppression (SFS) layer then converts the inferred unconstrained graph to the MST-based tree graph by a differentiable manner, thereby naturally incorporating the constraints into the graph generation. 

By integrating our feature suppression layer with a state-of-the-art transformer-based graph generator, we develop TreeFormer, which infers tree structures from images capturing plants. 
We evaluate the effectiveness of TreeFormer on different classes of plant images: Synthetic tree patterns, real-world root, and grapevine branch images. 
The results show that our constraint-aware graph generator accurately estimates the target tree structures compared to baselines.

\vspace{-4mm}
\paragraph{Contributions} 
Our contributions are twofold: First, we propose a novel method that tightly integrates learning-based graph generation methods with traditional graph algorithms using the newly-proposed SFS layer, which modifies intermediate features in the network, effectively mimicking the behavior of the non-differentiable graph algorithms. 
Second, building upon our constrained graph generation method, we develop TreeFormer, the first end-to-end method inferring skeletal structures from a single plant image, which benefits the agriculture and plant science field.

\section{Related Work}
We propose constraining the graph structures given by image-based graph generators, whose primary goal is plant skeleton estimation.
We, therefore, introduce the related work of plant skeleton estimation, graph generation from images, and constrained optimization for neural networks.

\subsection{Plant skeleton estimation}
Plant skeleton estimation is actively studied since it becomes a fundamental technique for downstream tasks related to plant phenotyping and cultivation~\cite{Okura3D}. 

\vspace{-3mm}
\paragraph{3D plant skeleton estimation}
Several methods are proposed to derive plant skeletons from 3D observations~\cite{Okura3D}. 
These methods often use point clouds acquired by LiDAR~\cite{skeleton3dTreeLiDAR,skeletonTreeLaserOld} or multi-view stereo (MVS)~\cite{MultiStereo3dPoint,skeleton3dMaizePoint}. Regardless of the 3D acquisition method, these works generally use a two-stage pipeline: Skeletonization~\cite{bucksch2014practical} followed by graph optimization using MST or Dijkstra's algorithm~\cite{AdTree,L1point,3dpointSeg,PointCloudSeg,StochasticSkeleton}, where the graph algorithms are required to convert a set of skeleton positions into a graph.

\vspace{-3mm}
\paragraph{2D plant skeleton estimation}
Compared to 3D methods, skeleton estimation from a single 2D image poses significant technical challenges due to the lack of depth information and severe occlusions despite the simplicity of data acquisition. Like 3D methods, existing 2D methods use a two-stage process involving skeletonization and graph optimization. To extract the skeleton regions on 2D images, plant region segmentation is often used for plants with relatively thin leaves~\cite{CenterDetection2D}.
Similarly, a neural network that converts an input image into a map representing 2D skeleton positions is used to mitigate the occlusions~\cite{isokane2018probabilistic}. To reason about the direction of intersecting branches, a recent work~\cite{Guyot} proposes to use vector fields representing branch direction instead of mask images, similar to the Part Affinity Fields (PAFs) used in OpenPose~\cite{OpenPose}. 

Unlike existing two-stage methods, we propose an end-to-end method that directly infers a tree graph representing plant skeletons in a single image.
Our experiments show that our end-to-end method achieves better accuracy than a recent two-stage method for 2D images.

\subsection{Graph generation from images}

Graph generation from images, sometimes called image-to-graph generation, is studied for extracting thin structures (\eg, road networks) or relations (\eg, scene graphs) from images~\cite{polygon-rnn++, GGT, Image2SMILES, SGTR, StrucLandDetect, PolyBuildSeg, PolyTransform, CurveGCN, Relationformer, Sat2Graph, SOTAWorkRNGDet, iCurbIL, RelTR_DETR}. 
Recent learning-based methods often use object detectors, which detect graph nodes (\eg, intersections in road networks) from images, and then aggregate the combinations of node features to predict the edges defined between two nodes as binary (\ie, existence of edges) or categorical (\eg, classification of edge relations) values. Some studies use external knowledge~\cite{2longtail,2recovering, Improving, glove} to improve the results.

Graph generators have usually taken autoregressive methods (\eg,~\cite{Molecule_auto_father, Molecule_auto1, Molecule_auto2, GGT, deeptree}) that output a graph by starting at an initial node and estimating neighboring graph nodes. 
Recent GNNs and transformers enable non-autoregressive graph generators~\cite{VAE_father, GraphVAE, PSG_query_matching, SGTR, MoFlow, RelTR_DETR} simultaneously estimating the entire graph. Autoregressive methods are prone to errors during the estimation process, and the state-of-the-art non-autoregressive method, RelationFormer~\cite{Relationformer}, performs better than autoregressive methods, especially for medium to large graphs.

A few recent studies consider graph generation with tree-graph constraints in a different context.
For the molecule structure estimation~\cite{SpanningTreeMolecules, TreeMolecules_father, TreeMolecules_1}, these methods assume autoregressive graph generation, making it hard to work with complex and relatively large graphs like botanical plants.

\subsection{Constrained optimization for neural networks}
Constrained optimization is crucial for machine learning. In particular, introducing constraints in neural networks has become a recent trend~\cite{kotary2021end}. 

\vspace{-3mm}
\paragraph{Designing differentiable layers}
The most direct way to introduce additional constraints to neural networks is to make the constraints differentiable.
In the continuous domain, it is known that a convex optimization can be implemented as a differentiable layer~\cite{agrawal2019differentiable}. 
However, the design of differentiable layers for combinatorial optimization poses a significant challenge due to the difficulty of differentiation. 
Wilder~\etal~\cite{wilder2019melding} propose a differentiable layer for linear programming (LP) problems using continuous relaxation. This method is extended to mixed integer linear programming (MILP)~\cite{ferber2020mipaal} by splitting the problem into multiple LPs. 
MST, which we want to use as constraints, is known to be transformed into the class of MILP~\cite{myung1995generalized,pop2020generalized}. However, using differentiable layers for these complex combinatorial problems requires exponential computation time~\cite{kotary2021end} to obtain the exact solution and is practically unrealistic.

\begin{figure*}[t]
	\centering
	\includegraphics[width=\linewidth]{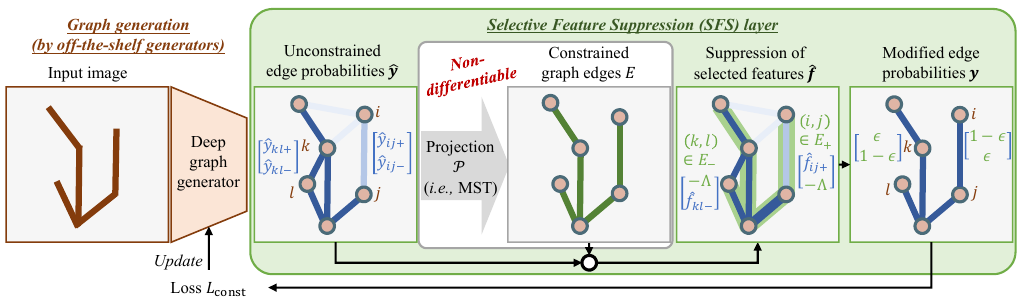}\vspace{-3mm}
	\caption{Overview of reparameterization layer that can be easily plugged into off-the-shelf graph generators. Given unconstrained edge predictions by graph generators, our method \emph{projects} it to the closest constrained graph (\ie, tree) using a non-differentiable MST algorithm. Comparing constrained and unconstrained edges, unwanted edge features are selectively suppressed so that the graph becomes the tree.}
	\label{fig:overview}
\end{figure*}

\vspace{-3mm}
\paragraph{Reparameterization for constrained optimization}
If the constraint function is difficult to differentiate, a simple alternative is to \emph{project} unconstrained inferences or model parameters into constrained space, which can be considered a use of reparameterization~\cite{kingma2013auto}. 

In gradient descent optimization, methods projecting unconstrained optimization parameters (\eg, model parameters in neural networks) to the closest ones satisfying the given constraint are called projected gradient descent (PGD). PGD is often used for traditional optimization problems, directly optimizing the input variables~\cite{PGD-old, PGD-CT}. While PGD can be used for neural network optimization, such as for generating adversarial examples~\cite{madry2018towards}, designing projection functions for neural networks is challenging. It requires mapping a large number of model parameters into a space satisfying complex constraints, where the constraints are often more naturally defined on the model output.

Instead of designing a projection function for the model parameters, the model's output can be projected onto the subspace that satisfies the constraints during the training loop. 
Since reparameterization for model output can be easily integrated with existing neural network models, they are often used for domain-specific applications such as coded aperture optimization with hardware constraint~\cite{CameraConstraints} and internal organ segmentation with given parametric shape models~\cite{SurveyMedicalConstraints, MedicalSegmentationConstraints}.
We take this approach in our SFS layer, easily plugging it into off-the-shelf end-to-end graph generation methods without preparing the differentiable implementation of the constraints (\ie, MST algorithm). 

\section{Tree-constrained Graph Generation}
We here describe the constrained graph generation method. \Fref{fig:overview} summarizes the proposed \REP layer, which casts the original unconstrained edge probabilities to the constrained domain.

\subsection{Problem statement}
We here consider a simple setting of neural-network-based graph generation, where the model outputs the prediction of the edge probabilities (\ie, the edge exists or not) defined for a pair of nodes, while this can be extended to a multi-class classification setting straightforwardly.

Our goal is to design a tree-constrained graph generator $\func$ that converts a given image $\img$ to a tree graph $\graph$ as
\begin{equation}
\label{eq:graph_const}
\graph = (\vertex,\edge) = \func(\img) \quad \mathrm{s.t.} \quad \edge \in \constraint,
\end{equation}
where the graph $\graph$ consists of a set of nodes (or objects) $\vertex$ and edges (or relations) $\edge$. Here, $\constraint$ denotes all possible edge patterns forming a tree graph given the set of nodes $\vertex$.

We consider a (non-differentiable) projection function $\Proj$ that maps an unconstrained graph predicted by graph generators to the constrained graph $G$. Let the edge probabilities defined between each node pairs as $\{\unconst{\V{y}}_{\ijsub}\}_{(i,j)\in{V\times V}}$.
\begin{equation}
\label{eq:probs}
\unconst{\V{y}}_{\ijsub}=[\unconst{y}^{+}_{\ijsub}, \unconst{y}^{-}_{\ijsub}]^\top \quad \mathrm{s.t.} \quad \|\unconst{\V{y}}_{\ijsub}\|_1=1,   
\end{equation}
in which $\unconst{y}^{+}_{\ijsub}$ and $\unconst{y}^{-}_{\ijsub}$ respectively denote the edge existence and non-existence probabilities.
The projection function $\Proj$ then read as
\begin{equation}
(\vertex,\edge) = \Proj_{\edge \in \constraint}(\vertex,\{\unconst{\V{y}}_{\ijsub}\}),
\label{eq:P_I2G}
\end{equation}
which are given by traditional graph algorithms with combinatorial optimization.
We assume the projection function $\Proj$ converts the existence probability (or category prediction) of graph edges while leaving the graph nodes $V$ unchanged.
A typical example of $\Proj$ can be designed using the MST algorithm we use in our TreeFormer implementation, which projects an arbitrary graph into a tree structure by modifying the existence of graph edges based on the costs defined between each pair of nodes.

We aim to develop a differentiable function $\Rep$ that mimics the non-differentiable projection $\Proj$. Plugging with the \emph{unconstrained} graph generator $\unconst{\func}$, \eref{eq:graph_const} is rewritten as
\begin{equation}
\begin{array}{ll}
\graph = (\vertex,\edge) = \Rep_{E \in \constraint} (\vertex,\{\unconst{\V{y}}_{\ijsub}\}), \\
\quad (\vertex,\{\unconst{\V{y}}_{\ijsub}\}) = \unconst{\func}(\img),
\end{array}
\label{eq:rep}
\end{equation}
where the whole process is differentiable.

\subsection{SFS layer}
Here, we describe an implementation of the SFS layer for constrained graph generation. As described in \eref{eq:probs}, the unconstrained graph generator $\unconst{\func}$ computes the probability of \emph{unconstrained} edge existence between the $i$-th and $j$-th nodes, $\unconst{\V{y}}_{\ijsub}=[\unconst{y}^{+}_{\ijsub}, \unconst{y}^{-}_{\ijsub}]^\top$.
In neural networks, $\unconst{\V{y}}_{\ijsub}$ is usually computed through the softmax activation $\softmax$ applied to the output feature vector of the final layer $\unconst{\V{f}}_{\ijsub}=[\unconst{f}^{+}_{\ijsub},\unconst{f}^{-}_{\ijsub}]^\top\in\real^2$ as
\begin{equation}
\unconst{\V{y}}_{\ijsub} = \softmax(\unconst{\V{f}}_{\ijsub}). 
\label{eq:feat}
\end{equation}
The set of unconstrained graph edges $\unconst{E}$ are then obtained by comparing the edge existence probabilities as
\begin{equation}
\label{eq:thresholding}
\unconst{\edge} = \{ (i, j) \mid \unconst{y}^{+}_{\ijsub} > \unconst{y}^{-}_{\ijsub} \},
\end{equation}
in which $\unconst{E}$ records node pairs where the edge exists. 

Suppose the projection function $\Proj$ converts the set of unconstrained edge probabilities $\{\unconst{\V{y}}_{\ijsub}\}$ to a set of constrained edges $E$. 
Let the difference of two sets be $E^+ = E - \unconst{E}$ and $E^- = \unconst{E} - E$, denoting the sets of edges newly added and removed by the projection.
To mimic discrete (and non-differentiable) inferences by $\Proj$ in differentiable end-to-end learning, we modify the edge features corresponding to $E^+ \cup E^-$ in the differentiable forward process. 

Specifically, what we want to get is the edge probabilities that approximate the constrained edges $E$, denoted as
\begin{equation}
\label{eq:naive}
\V{y}_{\ijsub} = \left\{
\begin{array}{l@{\hspace{2em}}l}
\left[1-\epsilon,\:\:\:\:\epsilon\:\:\:\: \right]^\top           & ((i,j)\in E^+)\\
\left[\:\:\:\:\epsilon\:\:\:\:,  1-\epsilon \right]^\top           & ((i,j)\in E^-)   \\
\left[\unconst{{y}}^{+}_{\ijsub},  \unconst{{y}}^{-}_{\ijsub} \right]^\top       & \mathrm{(otherwise)}.
\end{array}
\right.
\end{equation}
When $\epsilon$ is small enough, the constrained output $\V{y}_{\ijsub}$ perfectly mimics the output by the projection function $\Proj$. 
However, the direct modification of the edge probabilities naturally disconnects the computation graph. 
Therefore, we modify the unconstrained feature vector $\unconst{\V{f}}_{\ijsub}$ so that the corresponding edge probabilities $\V{y}_{\ijsub}$ follows \eref{eq:naive}. Specifically, since $\V{y}_{\ijsub}$ is computed through the softmax function $\softmax$, it is achieved via the following minimal modification that \emph{selectively} suppresses the feature values by replacing them with a constant\footnote{See the supplementary materials for the derivation.} as
\begin{equation}
\begin{array}{l@{\hspace{2em}}l}
f^{-}_{\ijsub} := -\Lambda & ((i,j) \in E^+) \\
f^{+}_{\ijsub} := -\Lambda & ((i,j) \in E^-),
\end{array}
\label{eq:rep2}
\end{equation}
where $\Lambda$ is assumed to be large enough to make $\exp(-\Lambda) \sim 0$. Given modified features $\V{f}_{\ijsub} = [f^{+}_{\ijsub},f^{-}_{\ijsub}]^\top$, the softmax activation $\softmax$ normalizes and converts them to edge probability $\V{y}_{\ijsub}$. 

In summary, from Eqs.~(\ref{eq:feat}) and (\ref{eq:rep2}), the constrained edge prediction between $i$-th and $j$-th nodes,~\hbox{$\V{y}_{ij} = [y_{ij+},y_{ij-}]^\top$}, is obtained as
\begin{equation}
\label{eq:rep_layer}
\V{y}_{\ijsub} = \left\{
\begin{array}{l@{\hspace{2em}}l}
\softmax( [\unconst{{f}}^{+}_{\ijsub}, \:\:-\Lambda\:]^\top)  \quad  & ((i,j)\in E^+)   \\
\softmax( [\:\:-\Lambda\:,      \unconst{{f}}^{-}_{\ijsub}]^\top)       & ((i,j)\in E^-)    \\
\softmax( [\unconst{{f}}^{+}_{\ijsub},  \unconst{{f}}^{-}_{\ijsub}]^\top)       & \mathrm{(otherwise)}.
\end{array}
\right.
\end{equation}
After the reparameterization, the set of edges computed from $\{\V{y}_{\ijsub}\}$ in the same way as \eref{eq:thresholding} is guaranteed to be equal to $E$ inferred by the discrete projection function $\Proj$ when $\Lambda$ is large enough. 

\subsection{Analysis}
The common auto differentiation libraries automatically compute the gradient of the SFS layer.
Although it can disconnect the computation path at a feature, since we keep at least one of the original features (either $\unconst{{f}}^{+}_{\ijsub}$ or $\unconst{{f}}^{-}_{\ijsub}$), the backpropagation path to the backbone graph generation network is not disconnected\footnote{This is akin to the dropout layer often used in neural networks.}.
Here, we briefly analyze the behavior of the SFS layer. The supplementary materials provide a detailed analysis, including mathematical proofs. 

When using the cross-entropy loss $\Loss_\text{CE}$ to evaluate the availability of the graph edges, the derivative to be backpropagated to the backbone graph generator is approximated as\footnote{We omit the subscript $\ijsub$ for simplicity.}
\begin{align}
\label{eq:rep_gradient}
\frac{\partial \Loss_\text{CE}}{\partial \unconst{\V{f}}} &\sim
\left\{
\begin{array}{l@{\hspace{2em}}l}
\left[\:\:1-t^{+}\:, \:\:\:\:\:\:0\:\:\:\:\:\:\:\right]^\top & \hspace{-3mm}((i,j)\in E^+)    \\
\left[\:\:\:\:\:\:\:0\:\:\:\:\:\:, \:\:1-t^{-}\:\right]^\top & \hspace{-3mm}((i,j)\in E^-)    \\
\left[y^{+}-t^{+}, y^{-}-t^{-}\right]^\top & \hspace{-3mm}(\mathrm{otherwise}),
\end{array}
\right.
\end{align}
where $\V{t}=[t^{+},t^{-}]^\top\in\{0,1\}^2$ denotes the ground truth edge existence and non-existence for the node pair $\ijsub$. 
Our method modifies the computation graph of the network when the MST algorithm disagrees with the output of graph generation model (\ie, $(i,j) \in E^+\cup E^-$), but in different ways for derivatives of each feature value $\frac{\partial \Loss_\text{CE}}{\partial \unconst{{f^+}}}$ or~$\frac{\partial \Loss_\text{CE}}{\partial \unconst{{f^-}}}$.

Without loss of generality, we consider the case when the MST algorithm \emph{adds} an edge, \ie, $(i,j) \in E^+$. 
When the MST \emph{correctly} modify the edge availability (\ie, $t^+=1$), the gradient vector becomes small, $\frac{\partial \Loss_\text{CE}}{\partial \unconst{\V{f}}} \sim \V{0}$, which is the behavior we expect.
On the other hand, if the MST \emph{incorrectly} adds the edge (\ie, $t^+=0$), the gradient becomes $[1,0]^\top$, which strongly penalizes the positive edge probability, where the norm of the gradient vector is always larger than unconstrained ones\footnote{See supplementary materials for the mathematical proof.}.
Therefore, the behavior of our simple reparameterization strategy is reasonable in practice.

\section{TreeFormer: A Plant Skeleton Estimator}
\label{sec:TreeFormer}

We develop TreeFormer, an implementation of the SFS layer to a state-of-the-art graph generator. 
This section first recaps the graph generator~\cite{Relationformer} and then details how we introduce tree structure constraint.

\subsection{RelationFormer: A brief recap}
RelationFormer~\cite{Relationformer} is the state-of-the-art non-autoregressive graph generation method. 
This method uses an end-to-end architecture that combines an object (node) detector and relation (edge) predictor, which shows superior performance for unconstrained graph generation.
The object detection part is based on deformable DETR~\cite{Deformable_DETR}, which is trained to extract graph nodes (\eg, objects) and global features from a given image. 
Specifically, given the extracted image features, the transformer decoder outputs a fixed number of object queries ([obj]-tokens) representing each of the nodes and a relation query ([rtn]-token) describing the global features, including node relations. \looseness=-1

The relation prediction head outputs the relationship (\ie, edge existence or category) from the detected pairs of objects (\ie, [obj]-tokens) and the global relation (\ie, [rtn]-tokens). This module is implemented as a multi-layer perceptron (MLP) headed by layer normalization~\cite{LayerNorm}. 
RelationFormer is trained using the sum of loss functions related to object detection and edge (relation) estimation, where edge (relation) loss $\Loss_\text{edge}$\footnote{Denoted as $\Loss_\text{rln}$ in the original paper~\cite{Relationformer}, we use $\Loss_\text{edge}$ for generality.} evaluates the edge existence or category between node pairs using cross-entropy loss.

\subsection{Tree-constrained graph generation}
To introduce the tree structure constraint, we use Kruskal's MST algorithm~\cite{Kruskal} implemented in NetworkX\footnote{\url{https://networkx.org/}, last accessed on July 15, 2024.}. To extract a tree from an unconstrained graph predicted by RelationFormer, we use the edge non-existence probabilities $\{\unconst{y}^{-}_{\ijsub}\}$ as the edge cost for the MST algorithm to span the tree on edges with higher existence probabilities.

We implement the SFS layer on top of the relation prediction head in the RelationFormer. 
Specifically, the output features from the MLP after layer normalization are regarded as unconstrained features $\{\V{\unconst{f}}\}$.
In our experiments, we use $\Lambda = 10$ during training, where~\hbox{$\exp(-\Lambda) = 4.5\times 10^{-5}$}. We show an ablation study changing $\Lambda$ in the supplementary materials.

\vspace{-3mm}
\paragraph{Loss function} Our SFS layer affects the evaluation of the edge loss $L_\text{edge}$ in the graph generator, while the computation of other loss functions, such as for node detection, remains the same as in the original implementation. 
Our implementation uses both loss functions for original (unconstrained) and constrained edges. Denoting the ground-truth edges as $E_\text{GT}=\{(i,j) \mid t^{+}_{\ijsub} > t^{-}_{\ijsub}\}$, where $\V{t}_{\ijsub} = [t^{+}_{\ijsub},t^{-}_{\ijsub}]^\top\in\{0,1\}^2$,
the loss function for edge availability $\Loss_\text{edge}$ is modified as follows
\begin{equation}
\Loss_\text{edge} = \underbrace{\sum_{(i,j)}\Loss_\text{CE}(\unconst{\V{y}}_{\ijsub},\V{t}_{\ijsub})}_{\Loss_\text{unconst}} + \underbrace{\sum_{(i,j)}\Loss_\text{CE}({\V{y}}_{\ijsub},\V{t}_{\ijsub})}_{\Loss_\text{const}},
\end{equation}
where $\Loss_\text{CE}$ denotes the cross-entropy loss.

\section{Experiments}
\label{sec:experiment}
To assess the effectiveness of the proposed method and TreeFormer implementation, we perform experiments using synthetic and real image datasets.

\subsection{Datasets}
\begin{figure}[t]
	\centering
        \includegraphics[width=\linewidth]{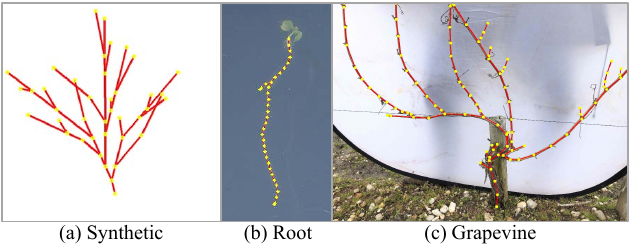} \vspace{-5mm}
	\caption{Example images from the dataset we used for our experiments. Annotated graphs are superimposed. Yellow dots and red lines indicate nodes and edges.}
 \vspace{-2mm}
	\label{fig:dataset_all}
\end{figure}

We use one synthetic and two real datasets, where examples are shown in \fref{fig:dataset_all}.
Supplementary materials describe the details of the datasets. 

\vspace{-3mm}
\paragraph{Synthetic dataset}
To systematically demonstrate the performance of our method, we perform an experiment using a large synthetic dataset.
We automatically generate images of tree patterns using pre-defined rules of Lindenmayer systems (L-system) \cite{L-system, L-system-papa}, which generate structural patterns using recursive processes.
We add randomness of branching patterns, branch length, and joint angles to increase the dataset variation.
The number of nodes in the graph is controlled at less than $100$. 
The resolution of the generated images is $512\times 512$ pixels.
We generated $100000$ images for training, $20000$ for validation, and $20000$ for testing.

\vspace{-3mm}
\paragraph{Root dataset}
We use photographs of early-growing roots of Arabidopsis, which are often important targets of analysis in plant science.
In this dataset, the graph structures are manually annotated.
The dataset contains $781$ root images, and we randomly divide them into $625$ training, $78$ validation, and $78$ test images. 
Each graph contains up to $117$ nodes. The image resolution is $570\times 190$ pixels.
We use data augmentation involving rotation, flipping, and cropping for the training dataset, which collectively expands the training dataset to $62,500$ images. 

\vspace{-3mm}
\paragraph{Grapevine dataset~\textnormal{\cite{Guyot}}} 
We use 3D2cut Single Guyot Dataset~\cite{Guyot} containing grapevine tree images captured in an agricultural field with annotated branch patterns. 
The dataset contains relatively complex structures; the graph contains up to $205$ nodes.
The resolution is $504\times 378$ pixels.
The dataset contains $1503$ images, and we use the dataset split the same as \cite{Guyot}, where $1185$ images are for training, and $63$ and $255$ images are for validation and testing, respectively.
We use data augmentation in the same manner as the root dataset, resulting in $118,500$ training images. 

\subsection{Evaluation metrics}
We use different metrics to capture spatial similarity alongside the topological similarity of the predicted graphs.

\paragraph{Street mover's distance (SMD)~\textnormal{\cite{GGT}}} SMD is a metric to assess the accuracy of the positions of graph edges, which is computed as the Wasserstein distance between the predicted and the ground truth edges. In our implementation, the distance is computed between densely sampled points on the edges, which is the same procedure as in the original paper proposing the SMD~\cite{GGT}. 

\vspace{-3mm}
\paragraph{TOPO score~\textnormal{\cite{he2018roadrunner}}} We compute the TOPO scores to evaluate the topological mismatch of the output graph. This metric consists of the precision, recall, and F1 scores of the graph nodes, which are evaluated considering the edge topology. We use the implementation used in Sat2Graph paper~\cite{Sat2Graph}, while we only evaluate the nodes with the degree $\neq 2$ that affect the tree structure, \ie, we only evaluate joint and leaf nodes in the graphs.

\vspace{-3mm}
\paragraph{Tree rate} To evaluate how well the output graph satisfies the constraint, we calculate the probability that the output graph forms a tree structure. While it is obvious that the tree rate becomes $100~\%$ for constrained methods, including ours, we are interested in how well the output of the unconstrained graph generation model can reflect the constraint by training on datasets that contain only tree graphs.

\subsection{Baselines}
Since the constrained graph generation task is new in this paper, there are few established baseline methods.
We compare our method with the state-of-the-art methods for 2D plant structure estimation and unconstrained graph generation.
Also, as an ablation study, we compare a simpler alternative to our method. Supplementary materials provide additional comparisons with other baseline methods, including autoregressive graph generation.

\vspace{-3mm}
\paragraph{Two-stage~\textnormal{\cite{Guyot}}} We implement a 2D plant skeleton estimation method based on a two-stage method involving skeletonization and graph optimization with reference to~\cite{Guyot}. Specifically, vector fields of branch directions are generated by a neural network, followed by graph optimization to generate branch structure, in which we find our implementation outperforms the naive re-implementation of the existing method~\cite{Guyot}. Specific implementations and analyses are described in the supplementary materials.

\vspace{-3mm}
\paragraph{Unconstrained~\textnormal{\cite{Relationformer}}} We compare the state-of-the-art (unconstrained) graph generation method, RelationFormer~\cite{Relationformer}. This method is identical to our method without applying the tree structure constraint.

\vspace{-3mm}
\paragraph{Test-time constraint} As a straightforward implementation of constrained graph generation, we apply MST only in the inference phase, where the graph generator is trained using the same procedure as the unconstrained method.

\begin{figure*}[t]
	\centering
	\includegraphics[width=\linewidth]{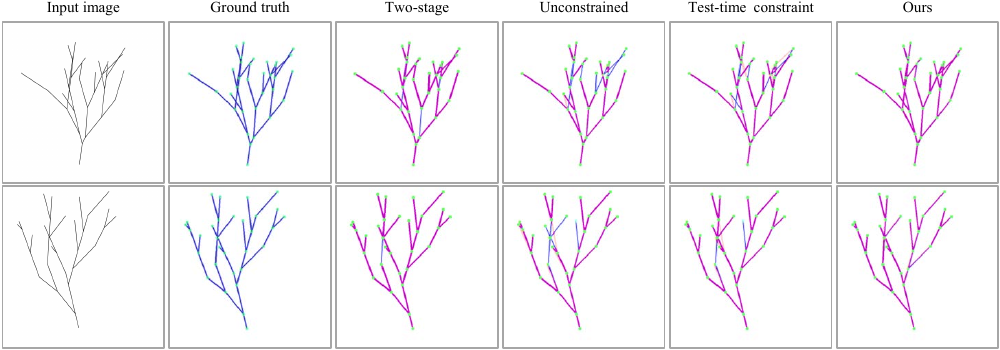}
\vspace{-5mm}
	\caption{Visual results for the synthetic tree pattern dataset. From left to right: Input images, results of the two-stage method (similar to~\cite{Guyot}), the unconstrained method (identical to RelationFormer~\cite{Relationformer}), a naive implementation with test-time constraint, and ours are shown. We translucently overlay the estimated and ground truth edges with red and blue lines, respectively. While all methods accurately detect nodes, only our method accurately predicts the availability of edges from given images compared to the baseline methods.}
\vspace{-3mm}
	\label{fig:synthetic}
\end{figure*}

\subsection{Implementation details}
For RelationFormer in our method and the baseline comparison, we use the official implementation\footnote{\url{https://github.com/suprosanna/relationformer}, last accessed on July 15, 2024.} on PyTorch.
For other hyperparameters, we follow the original RelationFormer implementation used for road network extraction. 
We used early stopping for all datasets and methods by selecting the model with the best validation performance and terminating training after $30$ epochs without improvement.
The training of our method takes approximately $141$ hours for the synthetic dataset, $10$ hours for the root dataset, and $98$ hours for the grapevine dataset, all conducted on eight NVIDIA RTX A100 GPUs.

\begin{figure*}[t!]
	\centering
    \subfloat[][Visual results for the root dataset.]{
	  \includegraphics[width=\linewidth]{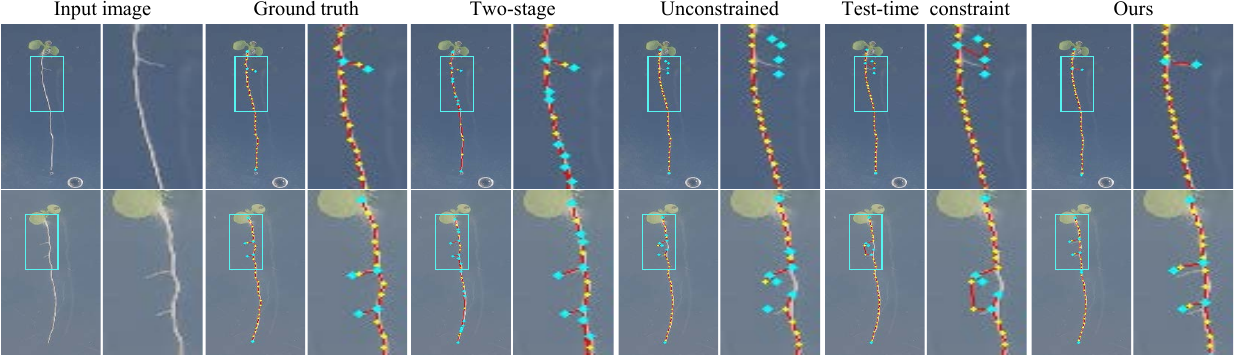}    
        \label{fig:root}
    }\\ \vspace{1mm}
    \subfloat[][Visual results for the grapevine dataset.]{
    	\includegraphics[width=\linewidth]{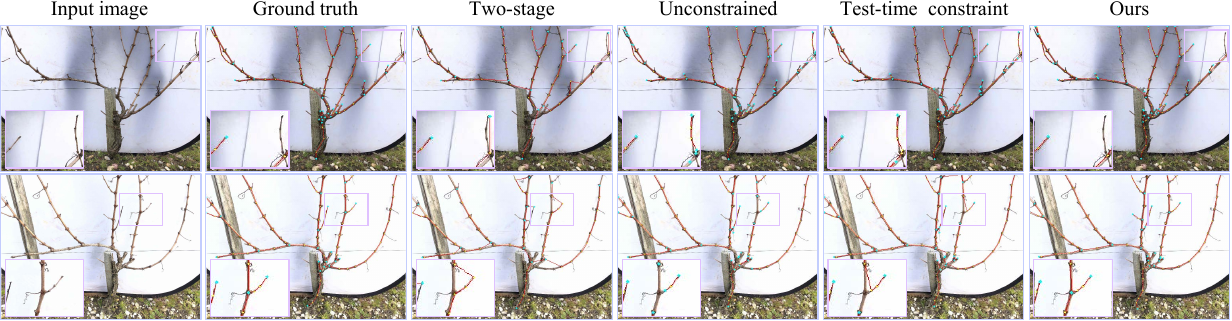}    
    	\label{fig:guyot}    
    }    
\vspace{-2mm}
	\caption{Visual results for the real image datasets. Red lines, yellow dots, and cyan dots indicate the predicted graph edges, nodes, and keypoints (\ie, joints and leaf nodes). Our method accurately estimates the target plant structures compared with baseline methods, demonstrating the applicability of our method for practical uses in plant science and agriculture.}
 \vspace{-3mm}
    \label{fig:real}
\end{figure*}

\begin{table}
\centering
\caption{Quantitative results. Our method significantly improves both the shape and topology of the predicted graph while enforcing the given constraints. The best scores are highlighted \textbf{bold}.}\vspace{-2mm}
\label{tab:tab1}
\resizebox{\linewidth}{!}{
\begin{tabular}{c|c|c|ccc|c}
\toprule
\multirow{2}{*}{Dataset}   & \multirow{2}{*}{Method}      & \multirow{2}{*}{SMD $\downarrow$} & \multicolumn{3}{c|}{TOPO score $\uparrow$} & Tree rate \\
                           &                                    &                      & Prec.     & Rec.     & F1       & [\%]   \\ \midrule
\multirow{4}{*}{Synthetic} & Two-stage~\cite{Guyot}             & $1.91 \times 10^{-3}$     & 0.940         & 0.886        & 0.912        &\textbf{100.0}     \\
                           & Unconstrained~\cite{Relationformer}& $1.43 \times 10^{-5}$     & 0.978         & 0.929        & 0.953        & 36.2      \\
                           & Test-time constraint               & $6.26 \times 10^{-6}$     & 0.977         & 0.953        & 0.965        &\textbf{100.0}     \\
                           & Ours                               & \bm{$4.78 \times 10^{-6}$}& \textbf{0.986}&\textbf{0.968}&\textbf{0.977}&\textbf{100.0}     \\ \midrule
\multirow{4}{*}{Root}      & Two-stage~\cite{Guyot}             & $4.83 \times 10^{-4}$     & 0.767         &0.732         &0.749         &\textbf{100.0}     \\
                           & Unconstrained~\cite{Relationformer}& $1.19 \times 10^{-4}$     & 0.831         & 0.633        & 0.719        & 35.9     \\
                           & Test-time constraint               & $1.52 \times 10^{-4}$     & 0.829         & 0.771        & 0.799        &\textbf{100.0}     \\
                           & Ours                               & \bm{$8.82 \times 10^{-5}$}&\textbf{0.861} &\textbf{0.807}&\textbf{0.833}&\textbf{100.0}     \\ \midrule
\multirow{4}{*}{Grapevine} & Two-stage~\cite{Guyot}             & $4.24 \times 10^{-4}$     & 0.677         & 0.589        & 0.630        &\textbf{100.0}     \\
                           & Unconstrained~\cite{Relationformer}& $1.45 \times 10^{-4}$     & \textbf{0.963}& 0.559        & 0.708        & 0.0      \\
                           & Test-time constraint               & $1.47 \times 10^{-4}$     & 0.896         & 0.840        & 0.867        &\textbf{100.0}     \\
                           & Ours                               & \bm{$1.03 \times 10^{-4}$}& 0.899         &\textbf{0.843}&\textbf{0.870}&\textbf{100.0}     \\ \bottomrule
\end{tabular}
}
\end{table}

\subsection{Results on synthetic dataset}
\Fref{fig:synthetic} shows visual results for the synthetic dataset, where the red and blue lines indicate the predicted and ground truth edges, respectively. 
Since they are shown translucently, if the estimated edge overlaps the true edge, it is displayed in purple. 
Similarly, cyan and yellow dots indicate the nodes, which merge into green if correctly estimated. From the results, all methods correctly estimate the node positions.
The existing unconstrained method outputs isolated edges and cycles. Although the two-stage and test-time constraint methods enforce the tree structure constraint, they often produce incorrect edges. 
Compared to the baselines, our method accurately generates the graph edges. 

The above trend can be quantitatively confirmed in \Tref{tab:tab1}. The unconstrained method produces tree structures with only about $30$~\% probability, even though all the training graphs form tree structures. Although introducing the test-time constraint and two-stage methods improves the shape and topology, there are still many incorrect estimates. Compared to those baseline methods, our method significantly improves both edge positions and graph topology.

\subsection{Results on real datasets}
\Fref{fig:real} show results of skeleton estimation for two real-world datasets. 
For these figures, red lines, yellow dots, and cyan dots indicate the edges, nodes, and keypoints (\ie, joints and leaf nodes), respectively.
In agreement with the synthetic results, our method predicts visually better structures, while the unconstrained model hardly produces tree structures.
The method with test-time constraint clearly produces false edges, as shown in the results for the grapevine images. 
The two-stage method is often sensitive to the node detection error, leading to unnecessary ({\it cf}.~\fref{fig:root}) or missing ({\it cf}.~\fref{fig:guyot}) keypoints.
In these practical settings, our end-to-end pipeline especially benefits from the simultaneous optimization of edge and node detection, resulting in faithful predictions at both nodes and edges for real-world datasets.

The quantitative results in \Tref{tab:tab1} confirm the advantage of our method for real-world scenes.
Our method shows particularly compelling results on grapevine datasets with relatively complex branching structures, outperforming the second-best method (test-time constraint) by approximately $30~\%$ improvement on the edge accuracy evaluated by SMD.

\begin{figure}
	\centering
        \includegraphics[width=\linewidth]{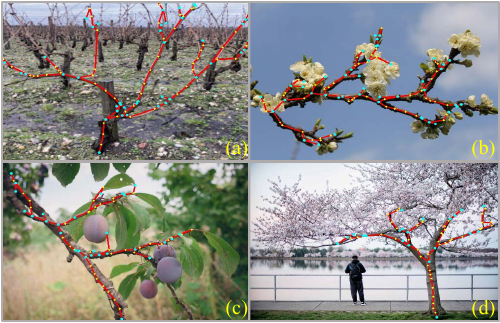}
        \vspace{-5mm}
	\caption{Results for the additional test images of (a) a grapevine tree under a natural background and (b--d) other tree species.}
        \vspace{-3mm}
	\label{fig:sakura}
\end{figure}

\vspace{-3mm}
\paragraph{Generalization ability}
We test our model on additional test datasets to validate the out-of-domain performance of our method, using the model trained on the Grapevine dataset.
Although the model is trained with grapevine trees with few background textures, it successfully works for grapevine images with background textures (\fref{fig:sakura}(a)) and for other tree species (\fref{fig:sakura}(b--d)).
These results highlight the generalizability of our method.

\section{Conclusion}
\vspace{-2mm}
We present the first attempt at tree-constrained graph generation from a single image, especially for plant skeleton estimation. We combine modern learning-based graph generators and traditional graph algorithms via the SFS layer, easily integrated with off-the-shelf graph generators.

\vspace{-4.5mm}
\paragraph{Limitations} 
We use graph algorithms during each training iteration, taking a longer training time than unconstrained methods, where fast GPU-based MST implementations (\eg,~\cite{vineet2009fast}) can improve computational performance.
The success of our method depends on the accuracy of the underlying graph generation model, as we see a few undetected nodes in the visual results.
Unlike \emph{universal} human skeleton estimation such as OpenPose~\cite{OpenPose}, our method requires domain-specific training due to the excessive variety of real-world plant appearances and structures, although we show certain generalizability in our experiments. 

\vspace{-4.5mm}
\paragraph{Acknowledgements} 
We thank Professor Yasuyuki Matsushita for insightful discussions throughout the study. We also thank Momoko Takagi, Manami Okazaki, and Professor Kei Hiruma for providing us with root images. 
This work was partly supported by JSPS KAKENHI Grant Numbers JP22K17910, JP23H05491, and JP21H03466, and JST FOREST Grant Number~\hbox{JPMJFR206F}.


\clearpage

\setcounter{section}{0}
\setcounter{figure}{0}
\setcounter{table}{0}
\setcounter{equation}{0}
\renewcommand\thesection{\Alph{section}}
\renewcommand\thefigure{S\arabic{figure}}
\renewcommand\thetable{S\arabic{table}}
\renewcommand\theequation{S\arabic{equation}}

\maketitlesupplementary

This supplementary material provides additional information, including details of our SFS layer (\sref{sec:supp_analysis}), dataset details (\sref{sec:supp_dataset}), implementation details of the baseline methods (\sref{sec:supp_baseline}), performance analysis of our method (\sref{sec:performance_analysis}), other design choices (\sref{sec:supp_ablation}), and more visual results (\sref{sec:supp_results}).

\section{Details of SFS layer}
\label{sec:supp_analysis}

\subsection{Motivation}
Our method infers a tree graph via MST as formulated in Eq.~(1)--Eq.~(4) in the main paper. Our task's goal is to optimize the graph generation network so that the \textbf{final output (\ie, tree graph via MST) becomes similar to the ground-truth tree graph}. 
Since MST modifies the edge availability in \emph{unconstrained} inferences, the unconstrained methods evaluating the unconstrained graph edges are indirect. 
Instead, our method \textbf{directly evaluates the quality of the final output tree} by mimicking MST. While experiments highlight our method's benefit, the following theoretical analysis also supports this intuition.

\subsection{Derivation}
This section details the derivation of Eq.~(10) in the main paper. To make this material self-contained, we repeat several descriptions in the main paper.

As discussed in the main paper, we consider the edge probabilities $\unconst{\V{y}}_{\ijsub} =[\unconst{y}^{+}_{\ijsub},\unconst{y}^{-}_{\ijsub}]^\top$ is usually computed through the softmax activation $\softmax$ applied to the output feature vector of the final layer $\unconst{\V{f}}_{\ijsub}=[\unconst{f}^{+}_{\ijsub},\unconst{f}^{-}_{\ijsub}]^\top$ as

\begin{align}
\begin{array}{ll}
\unconst{\V{y}}_{\ijsub} = \softmax(\unconst{\V{f}}_{\ijsub}) \\
= \Bigg[\frac{\exp(\unconst{f}^{+}_{\ijsub})}{\exp(\unconst{f}^{+}_{\ijsub})+\exp(\unconst{f}^{-}_{\ijsub})},
\frac{\exp(\unconst{f}^{-}_{\ijsub})}{\exp(\unconst{f}^{+}_{\ijsub})+\exp(\unconst{f}^{-}_{\ijsub})}\Bigg]^\top.
\end{array}
\label{eq:feat_softmax}
\end{align}

The set of unconstrained graph edges $\unconst{E}$ are then obtained by comparing the edge existence probabilities as
\begin{equation}
\label{eq:thresholding}
\unconst{\edge} = \{ (i, j) \mid \unconst{y}^{+}_{\ijsub} > \unconst{y}^{-}_{\ijsub} \},
\end{equation}
in which $\unconst{E}$ records node pairs where the edge exists. 

Suppose the projection function $\Proj$ converts the set of unconstrained edge probabilities $\{\unconst{\V{y}}_{\ijsub}\}$ to a set of constrained edges $E$. 
Let the difference of two sets be $E^+ = E - \unconst{E}$ and $E^- = \unconst{E} - E$, denoting the sets of edges newly added and removed by the projection.
To mimic the discrete (and non-differentiable) inferences by $\Proj$ in the differentiable end-to-end learning, we modify the edge features corresponding to $E^+ \cup E^-$ in the differentiable forward process. 
Here, we want to get the edge probabilities that approximate the constrained edges $E$, which can be denoted as
\begin{align}
\V{y}_{\ijsub} 
= \left\{
\begin{array}{l@{\hspace{2em}}l}
\label{eq:hard}
\left[\:\:\:\:1\:\:\:\:,\:\:\:\:0\:\:\:\: \right]^\top           & ((i,j)\in E^+)\\
\left[\:\:\:\:0\:\:\:\:,\:\:\:\:1\:\:\:\: \right]^\top           & ((i,j)\in E^-)   \\
\left[\unconst{{y}}^{+}_{\ijsub},  \unconst{{y}}^{-}_{\ijsub} \right]^\top       & \mathrm{(otherwise)}.
\end{array}
\right. \\
\label{eq:epsilon}
\sim \left\{
\begin{array}{l@{\hspace{2em}}l}
\left[1-\epsilon,\:\:\:\:\epsilon\:\:\:\: \right]^\top           & ((i,j)\in E^+)\\
\left[\:\:\:\:\epsilon\:\:\:\:,  1-\epsilon \right]^\top           & ((i,j)\in E^-)   \\
\left[\unconst{{y}}^{+}_{\ijsub},  \unconst{{y}}^{-}_{\ijsub} \right]^\top       & \mathrm{(otherwise)}.
\end{array}
\right.
\end{align}
When $\epsilon$ is small enough, the constrained output $\V{y}_{\ijsub}$ perfectly mimics the output by the projection function $\Proj$. Our goal is to modify the feature vector $\unconst{\V{f}}_{\ijsub}$ so that it makes the probabilities as \eref{eq:epsilon} through the softmax activation.

In the SFS layer, we replace the features as 
\begin{equation}
\begin{array}{l@{\hspace{2em}}l}
f^{-}_{\ijsub} := -\Lambda & ((i,j) \in E^+) \\
f^{+}_{\ijsub} := -\Lambda & ((i,j) \in E^-),
\end{array}
\label{eq:rep2}
\end{equation}
where $\Lambda$ is assumed to be large enough. Given modified features $\V{f}_{\ijsub} = [f^{+}_{\ijsub},f^{-}_{\ijsub}]^\top$, the softmax activation $\softmax$ normalizes and converts them to edge probability $\V{y}_{\ijsub}$ as
\begin{equation}
\label{eq:rep_layer}
\V{y}_{\ijsub} = \left\{
\begin{array}{l@{\hspace{2em}}l}
\softmax( [\unconst{{f}}^{+}_{\ijsub}, \:\:-\Lambda\:]^\top)  \quad  & ((i,j)\in E^+)   \\
\softmax( [\:\:-\Lambda\:,      \unconst{{f}}^{-}_{\ijsub}]^\top)       & ((i,j)\in E^-)    \\
\softmax( [\unconst{{f}}^{+}_{\ijsub},  \unconst{{f}}^{-}_{\ijsub}]^\top)       & \mathrm{(otherwise)}.
\end{array}
\right.
\end{equation}

Without loss of generality, we discuss the case in $(i,j)\in E^+$. Substituting \eref{eq:rep_layer} into \eref{eq:feat_softmax} yields
\begin{align}
\begin{array}{ll}
  \V{y}_{\ijsub} &=\left[\frac{\exp(\unconst{f}^{+}_{\ijsub})}{\exp(\unconst{f}^{+}_{\ijsub})+\exp(-\Lambda)},
\frac{\exp(-\Lambda)}{\exp(\unconst{f}^{+}_{\ijsub})+\exp(-\Lambda)}\right]^\top \nonumber \\
&= \left[\frac{\exp(\unconst{f}^{+}_{\ijsub})}{\exp(\unconst{f}^{+}_{\ijsub})+\epsilon'},
\frac{\epsilon'}{\exp(\unconst{f}^{+}_{\ijsub})+\epsilon'}\right]^\top
\quad  ((i,j)\in E^+),
\end{array}
\end{align}
where $\epsilon'=\exp(-\Lambda)\sim 0$ when $\Lambda$ is large enough, leading to $\V{y}_{\ijsub}=[1-\epsilon, \epsilon]^\top$ as in \eref{eq:epsilon} by denoting $\epsilon = \frac{\epsilon'}{\exp(\unconst{f}^{+}_{\ijsub})+\epsilon'}\sim 0$.

\subsection{Detailed analysis}

We describe a detailed analysis of our reparameterization layer. As described in \eqref{eq:rep_layer}, the unconstrained edge feature between $i$ and $j$-th nodes $\unconst{\V{f}}_{\ijsub} = [\unconst{{f}}^{+}_{\ijsub},\unconst{{f}}^{-}_{\ijsub}]^\top$ is converted to constrained prediction of the edge availability $\V{y}_{\ijsub} = [y_{\ijsub}^+,y_{\ijsub}^-]^\top$
by selectively suppressing unwanted feature values.

When using the cross-entropy loss $\Loss_\text{CE}$ to evaluate the availability of the graph edges, the derivative to be backpropagated to the backbone graph generator is \footnote{We omit the subscript $\ijsub$ for simplicity.}
\begin{align}
\label{eq:rep_gradient}
\frac{\partial \Loss_\text{CE}}{\partial \unconst{\V{f}}} &= 
\left\{
\begin{array}{l@{\hspace{2em}}l}
\left[(1-\epsilon)-t^{+}, \:\:\:\:\:\:\:\:\:\:0\:\:\:\:\:\:\:\:\:\:\:\right]^\top & ((i,j)\in E^+)    \\
\left[\:\:\:\:\:\:\:\:\:\:0\:\:\:\:\:\:\:\:\:\:\:, (1-\epsilon)-t^{-}\right]^\top & ((i,j)\in E^-)    \\
\left[\:\:\:y^{+}-t^{+}\:\:\:\:\:, \:\:\:y^{-}-t^{-}\:\:\:\:\:\right]^\top & (\mathrm{otherwise}),
\end{array}
\right.\\
\label{eq:rep_gradient2}
&\sim
\left\{
\begin{array}{l@{\hspace{2em}}l}
\left[\:\:1-t^{+}\:, \:\:\:\:\:\:0\:\:\:\:\:\:\:\right]^\top & ((i,j)\in E^+)    \\
\left[\:\:\:\:\:\:\:0\:\:\:\:\:\:, \:\:1-t^{-}\:\right]^\top & ((i,j)\in E^-)    \\
\left[y^{+}-t^{+}, y^{-}-t^{-}\right]^\top & (\mathrm{otherwise}),
\end{array}
\right.
\end{align}
where $\V{t}=[t^{+},t^{-}]^\top$ denotes the ground truth edge existence and non-existence for the node pair $\ijsub$. 
Our method modifies the computation graph of the network when the MST algorithm does not agree with the output of the graph generation model (\ie, $(i,j) \in E^+\cup E^-$), but in different ways for derivatives of each feature value $\frac{\partial \Loss_\text{CE}}{\partial \unconst{{f^+}}}$ or $\frac{\partial \Loss_\text{CE}}{\partial \unconst{{f^-}}}$.

\begin{table}[t]
\centering
\caption{Case-by-case analyses of our reparameterization layer. For the columns of unconstrained features $\unconst{\V{f}}$, constrained prediction $\V{y}$, and the ground truth edge availability $\V{t}$, the table shows the index of the larger element. For example, the column $\unconst{\V{f}}$ will be $+$ when the edge feature for the positive edge availability is larger, \ie, $\unconst{{f}^+}>\unconst{{f}^-}$. The column $(i,j)$ displays $E^+$ or $E^-$ if the MST algorithm modifies the edge availability (in which the rows are also highlighted). For the remaining columns, $\uparrow$ and $\downarrow$ denote each value becoming (relatively) large or small, respectively.}
\label{tab:supp_analysis}
\centering
\resizebox{\linewidth}{!}{
    \begin{tabular}{c|ccc|c|c|cccc|c}
    \toprule
        \multirow{2}{*}{Case}& \multicolumn{3}{c|}{Feats \& probs} & GT & Loss &  \multicolumn{4}{c|}{Approx. derivatives} & \multirow{2}{*}{Descriptions} \\ 
        & $\unconst{\V{f}}$ & $(i,j)$ & $\V{y}$ & $\V{t}$ &  $\Loss_\text{CE}$ 
        & $\frac{\partial \Loss_\text{CE}}{\partial \unconst{f}^{+}}$ 
        & $\frac{\partial \Loss_\text{CE}}{\partial \unconst{f}^{-}}$
        & $\left|\frac{\partial \Loss_\text{CE}}{\partial \unconst{f}^{+}}\right|$ 
        & $\left|\frac{\partial \Loss_\text{CE}}{\partial \unconst{f}^{-}}\right|$\\ 
        \midrule
       1 & $+$    &       & $+$ & $[1,0]^\top$ & $\downarrow$  &$y^+-1$&$y^-$& $\downarrow$  & $\downarrow$ & Unmodified\\
       2 & $+$    &       & $+$ & $[0,1]^\top$ & $\uparrow$    &$y^+$&$y^--1$& $\uparrow$    & $\uparrow$  & Unmodified \\ 
\rowcolor{lightgray}
       3 & $+$    & $E^-$ & $-$ & $[1,0]^\top$ & $\uparrow$    &$0$&$1$&  $\downarrow$ & $\uparrow$ & MST \emph{incorrectly} modified\\ 
\rowcolor{lightgray}
       4 & $+$    & $E^-$ & $-$ & $[0,1]^\top$ & $\downarrow$  &$0$&$0$&  $\downarrow$ & $\downarrow$ & MST \emph{correctly} modified\\ 
\midrule                                
       5 & $-$    &       & $-$ & $[1,0]^\top$ & $\uparrow$    &$y^+-1$&$y^-$&  $\uparrow$   & $\uparrow$ &Unmodified \\ 
       6 & $-$    &       & $-$ & $[0,1]^\top$ & $\downarrow$  &$y^+$&$y^--1$&  $\downarrow$ & $\downarrow$ &Unmodified\\ 
\rowcolor{lightgray}
       7 & $-$    & $E^+$ & $+$ & $[1,0]^\top$ & $\downarrow$  &$0$&$0$&  $\downarrow$ & $\downarrow$ & MST \emph{correctly} modified\\
\rowcolor{lightgray}
       8 & $-$    & $E^+$ & $+$ & $[0,1]^\top$ & $\uparrow$    &$1$&$0$& $\uparrow$    & $\downarrow$ & MST \emph{incorrectly} modified\\ 
    \bottomrule
    \end{tabular}
}
\end{table}

\Tref{tab:supp_analysis} summarizes the case-by-case behavior, in which we can categorize the behaviors of the SFS layer into eight cases. Hereafter, we use $E^* \triangleq E^+\cup E^-$. 

\vspace{2mm}\noindent\textit{Case $(i,j)\notin E^*$ (Cases 1, 2, 5, 6)\quad}
When the graph algorithm (\ie, MST) does not modify the edge availability (\ie, cases 1, 2, 5, and 6 in the table), the behavior is the same as the usual cross-entropy loss for unconstrained edges. 

\vspace{2mm}\noindent\textit{Case $(i,j)\in E^* \;\; \& \;\; \V{y}\sim \V{t}$ (Cases 4, 7)\quad}
In these cases, the MST algorithm \emph{correctly} suppresses the unwanted features, where the constrained prediction $\V{y}$ becomes the approximation of the ground-truth edge availability $\V{t}$. The loss value becomes small, and the derivative is $\frac{\partial \Loss_\text{CE}}{\partial \unconst{\V{f}}} \sim \V{0}$. This is natural since our constrained graph generator produces correct predictions.

\vspace{2mm}\noindent\textit{Case $(i,j)\in E^* \;\; \& \;\; \V{y}\nsim \V{t}$ (Cases 3, 8)\quad}
In these cases, MST \emph{incorrectly} modifies the edge availability, \ie, the node pair $(i,j)$ belongs to $E^+$ or $E^-$, but the constrained prediction $\V{y}$ does not fit the ground truth $\V{t}$. 
Here, we mathematically discuss the behavior in these cases. Without loss of generality, we focus on Case 3, where MST \emph{incorrectly} removes an edge and compares the methods with and without tree-graph constraints using the SFS layer.

\paragraph{Case 3 (MST \emph{incorrectly} removes an edge)}
The following discussions can be straightforwardly extended to Case 8, where MST \emph{incorrectly} adds an edge.

\paragraph{Conditions:}
\begin{itemize}
\vspace{-2mm}
    \item Unconstrained prediction (edge exists): $\unconst{f}^+ > \unconst{f}^-$, 
    \item MST \emph{removes} the edge: $(i,j)\in E^-$,
    \item GT edge availability (edge exists): $[t^+,t^-] = [1,0]$.
\end{itemize}

\paragraph{Unconstrained method (without SFS layer)}
The gradient at $\unconst{f}^+$ by the unconstrained method is 
\begin{align}
    \frac{\partial \Loss_\text{unconst}}{\partial \unconst{f}^{+}} = \frac{\exp(\unconst{f}^+)}{\exp(\unconst{f}^+)+\exp(\unconst{f}^-)}-1.
\end{align}
Since $\unconst{f}^+>\unconst{f}^-$, it takes the value in the range of $(-0.5,0)$. 
Similarly,
\begin{align}
    \frac{\partial \Loss_\text{unconst}}{\partial \unconst{f}^{-}} = \frac{\exp(f^-)}{\exp(f^-)+\exp(f^+)}-0,
\end{align} 
thus the gradient at $\unconst{f}^-$ is inside $(0, 0.5)$. 
Thus, the gradient vector $\frac{\partial \Loss_\text{unconst}}{\partial \unconst{\V{f}}}$ is always shorter than $[-0.5,0.5]^\top$ (corresponding to the special case $\unconst{f}^+=\unconst{f}^-$).

\paragraph{Constrained method (Ours)}
As described in Eqs.~\eqref{eq:rep_gradient} and \eqref{eq:rep_gradient2}, our method yields the gradient as 
\begin{align}
    \frac{\partial \Loss_\text{const}}{\partial \unconst{f}^{+}} = 0, \qquad
    \frac{\partial \Loss_\text{const}}{\partial \unconst{f}^{-}} = 1-\epsilon \sim 1, \nonumber
\end{align}
\ie, $\frac{\partial \Loss_\text{const}}{\partial \unconst{\V{f}}}\sim [0,1]^\top$.

\paragraph{Comparisons}
While both methods control the features to increase the edge availability, the relation of gradient vectors $\|\frac{\partial \Loss_\text{const}}{\partial \unconst{\V{f}}}\| > \|\frac{\partial \Loss_\text{unconst}}{\partial \unconst{\V{f}}}\|$ always holds, which means our method strongly penalizes the incorrect estimates by MST by directly comparing the final estimation (\ie, tree graph) with the ground-truth edge availability, which highlights our key motivation---a \textbf{direct} control of the tree-constrained graph generation.

\section{Dataset Details}
\label{sec:supp_dataset}
We describe the details of the datasets used in our experiment. 

\paragraph{Synthetic tree pattern dataset}
To prepare the synthetic dataset, we implement a generator of two-dimensional tree patterns based on the L-system~\cite{L-system-papa}, a formal language for describing the growth of the structural form. 
The L-system recursively applies rewriting rules to the current structure to simulate the growth of branching structures.

\begin{figure}[t]
	\centering
	\includegraphics[width=\linewidth]{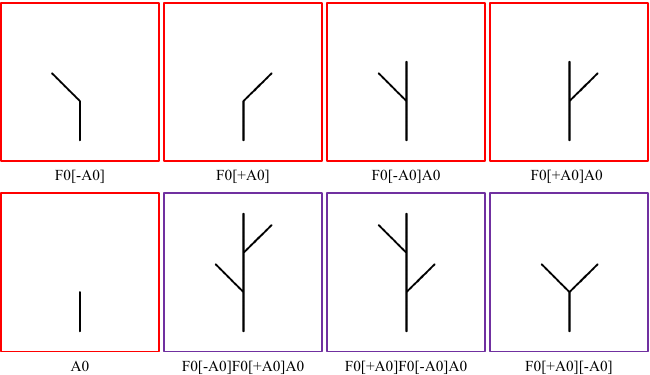}
	\caption{Atomic structures used for synthetic dataset generation. Three pre-defined initial structures are highlighted in purple. Eight pre-defined rewriting rules are used during the generation.}
	\label{fig:Atomic_structures}
\end{figure}

\begin{figure}[t]
	\centering
	\includegraphics[width=\linewidth]{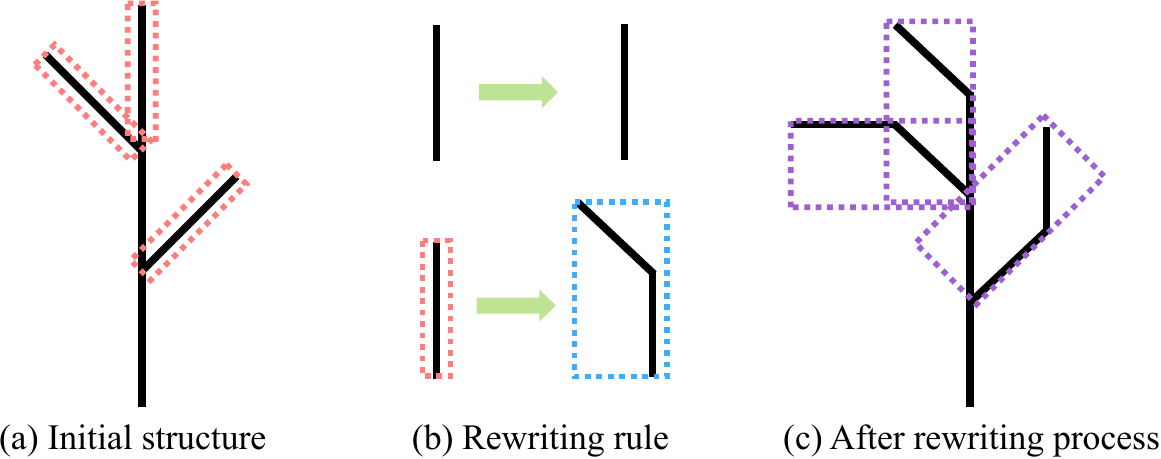}
	\caption{An example of the rewriting process. Suppose the initial structure is represented as \textbf{F0[+A0]F0[-A0]A0}. If a rewrite rule \textbf{F$\rightarrow$F{}; A{}$\rightarrow$F{}[-A{}]} is applied, \ie, \textbf{F} remains unchanged and \textbf{A} becomes \textbf{F{}[-A{}]}, the result of the rewrite process is \textbf{F0[+F1[-A1]]F0[-F1[-A1]]F1[-A1]}. The digits in the sequences indicate the number of times the rewrite is applied.}
	\label{fig:rewrite}
\end{figure}

\Fref{fig:Atomic_structures} shows the initial structures and the rewriting rules we used. At the beginning of the tree generation, an initial sequence is randomly chosen from the pre-defined sequences marked with a purple frame in the figure. At each iteration during the tree generation, the leaf edges (``\textbf{A}'' in the sequences) are replaced by a randomly chosen pattern from eight pre-defined ones. A simple example is shown in \fref{fig:rewrite}. We iterate the rewriting process a maximum of three times to generate a tree pattern.
We also add randomness to the branch length and joint angles in our dataset. We randomly choose a branch length of scaling $[0.5, 2.5]$ and joint angles of $[10^\circ,35^\circ]$. 

\paragraph{Root dataset}
\label{sec:root-augmentataion}
For the root dataset, the structure of the early-growing roots of Arabidopsis is manually annotated.
The structures are annotated by placing points (\ie, graph nodes) on the root path, where the distance between neighboring points may vary depending on the annotator and the images. We, therefore, resample the graph nodes with the same intervals. Starting from keypoints with the degree $\neq 2$ (\ie, joints and leaf nodes), we sample nodes at intervals of $8$ pixels along continuous branch segments.

For data augmentation, we apply flipping, rotation, cropping, noise, lighting, and scaling on the original images.
Supposing the roots are almost aligned at seeding, we limit the range of rotation angles in $[-9^\circ,+9^\circ]$.

\begin{table*}[tp]
\centering
\caption{Quantitative comparisons between our re-implementation of \cite{Guyot} and our two-stage baseline implementation. }
\label{tab:vinet}
\resizebox{\linewidth}{!}{
\begin{tabular}{c|c|ccc|c@{\hspace{2mm}}c}
\hline
\multirow{2}{*}{Method} & \multirow{2}{*}{SMD $\downarrow$} & \multicolumn{3}{c|}{TOPO score $\uparrow$} & \multicolumn{2}{c}{MSE $\downarrow$}\\
                      &                   & Prec.      & Rec.       & F1         & Node confidence  & Edge direction \\ \hline
Re-implementation of \cite{Guyot}    & $3.84\times 10^{-3}$      & $0.459$      & $0.365$      & $0.406$      & $6.95\times 10^{-3}$      & $1.01\times 10^{-2}$    \\ 
Our implementation of two-stage method            & $\bm{4.24\times 10^{-4}}$ & $\bm{0.677}$ & $\bm{0.589}$ & $\bm{0.630}$ & \bm{$1.19\times 10^{-3}$} & $\bm{2.67\times 10^{-3}}$   \\ \hline
\end{tabular}
}
\end{table*}

\begin{figure*}[tp]
	\centering
	\includegraphics[width=\linewidth]{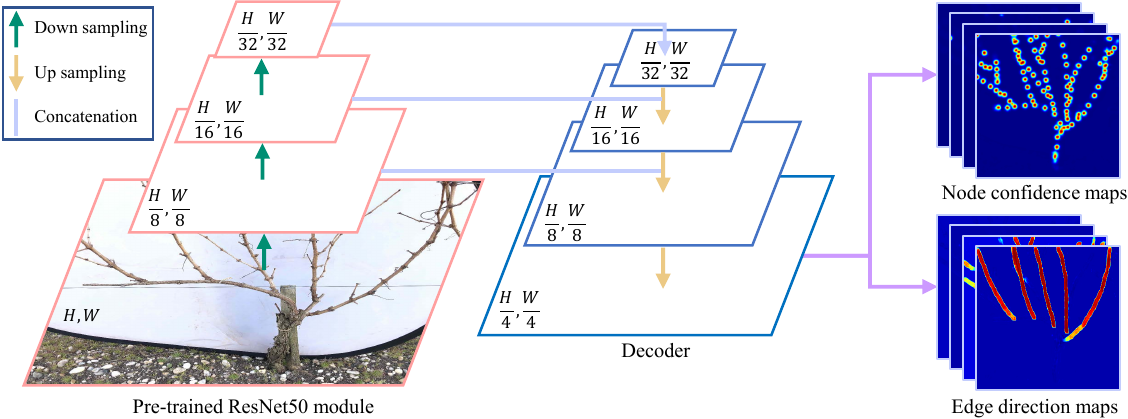}
	\caption{Network architecture for the first (skeletonization) stage of our two-stage baseline method.} 
	\label{fig:supp_two_stage_arch}\vspace{-2mm}
\end{figure*}

\begin{figure*}[tp]
	\centering
	\includegraphics[width=\linewidth]{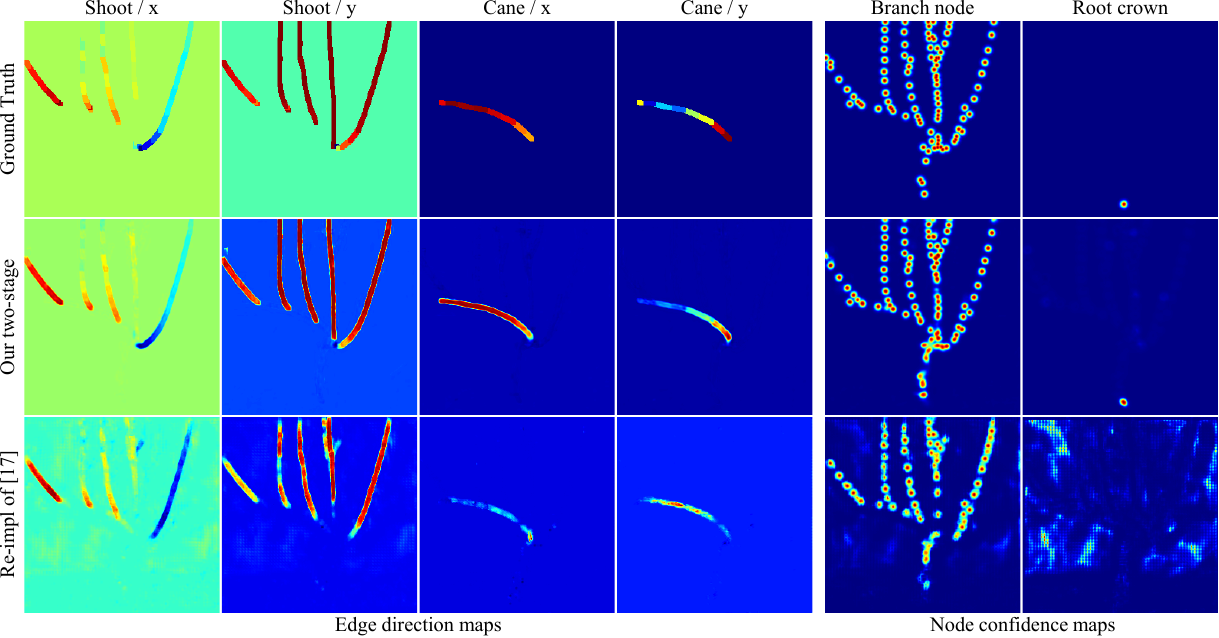}
	\caption{Visual comparisons between a re-implementation of \cite{Guyot} and our two-stage baseline implementation. Our implementation yields better node confidence and edge direction maps, which are the outputs of the first stage of these methods.}
	\label{fig:PAF_conf}
\end{figure*}

\begin{table*}[t]
\centering
\caption{Parameters used for the two-stage baseline method. $d$ denotes the distance threshold for the local maximum value search (\ie, non-maximum suppression) of node candidates. $\tau_m$ and $\tau_n$ are used as the thresholds for node detection from the confidence maps. For the detailed definitions, refer to the original paper~\cite{Guyot}.}
\label{tab:supp_params}
\resizebox{\linewidth}{!}{
\begin{tabular}{c|c|c|c|c|c|p{20mm}c}
\hline
\multirow{2}{*}{Dataset} & Image size   & Map size     & Node confidence & Edge direction & Node search distance & \multicolumn{2}{c}{Thresholds for node detection}  \\
                         & (W, H) [px]  & (W, H) [px]  &  diameter [px]  &  width [px]          & $d$ [px]    &  \centering $\tau_m$ & $\tau_n$        \\ \hline
Synthetic                & 512, 512     & 512, 512     & 4               & 5                    & 9     & \centering 0.97                   & 0.5         \\ 
Root                     & 570, 190     & 570, 190     & 3               & 5                    & 7     &\centering 0.99                   & 0.3         \\ 
Grapevine                & 1008, 756    & 256, 256     & 3               & 10                   & 25    &\centering 0.97                   & 0.5         \\ \hline
\end{tabular}
}
\end{table*}

\paragraph{Grapevine dataset}
\label{sec:grapevine-augmentataion}
We use 3D2cut Single Guyot Dataset~\cite{Guyot} containing manual annotations on branch structures. We perform data augmentation with rotation angles in $[-15^\circ, +15^\circ]$ in the same manner as~\cite{Guyot}.
This dataset also contains the classification of nodes (four classes) and edges (five classes) related to biological meanings. Since the existing two-stage method~\cite{Guyot} estimates these categories, we follow the same setup for the two-stage baseline method (refer to the next section for detailed discussions). For the other methods, including our TreeFormer implementation, we use only the binary class information (\ie, branch availability) for generalizability.

\section{Details of Baseline Methods}
\label{sec:supp_baseline}
We describe the implementation details for the baseline methods: The \emph{two-stage} method and the method with the \emph{test-time constraint}. Note the implementation for the other baseline, the \emph{unconstrained} method, is identical to the original RelationFormer~\cite{Relationformer}.

\subsection{Two-stage baseline}
\label{sec:supp_two_stage}
Our experiment implements a two-stage baseline involving skeletonization and graph optimization. This baseline implementation is based on ViNet~\cite{Guyot}, a state-of-the-art plant skeleton estimation method. Since the implementation of \cite{Guyot} is not publicly available, we re-implement the method with reference to the descriptions in the paper. Through the re-implementation, we find room for improvement in the two-stage baseline method.
\Tref{tab:vinet} compares the performance of our two-stage implementation with a naive re-implementation of \cite{Guyot}. The SMD and TOPO scores are the same metrics used in the main paper, and we also compare the mean squared error (MSE) of the first-stage output of both methods. Our implementation achieves a better performance; thus, we use the improved version for our experiment.
In the following, we describe the implementation details.

\paragraph{First stage: Skeletonization}
Similar to ViNet~\cite{Guyot}, the first stage of our implementation outputs the prediction of node and edge positions as single-channel confidence maps and two-channel vector fields (hereafter referred to as node confidence maps and edge direction maps, respectively). This step is similar to a widespread human pose estimation method \ie, OpenPose~\cite{OpenPose}, which jointly estimates the confidence of person keypoints and the Part Affinity Fields (\ie, two-channel vector fields).

While ViNet~\cite{Guyot} uses a sequence of residual blocks followed by the Stacked Hourglass Network~\cite{supp_stacked} for this stage, we use a pre-trained ResNet50~\cite{ResNet} for image feature extraction. This is for a fair comparison to our TreeFormer implementation, which also uses ResNet50 as the backbone\footnote{ResNet50 is actually used as the node detection module in RelationFormer (that is based on Deformable DETR~\cite{Deformable_DETR}), which is the basis of our TreeFormer, and we inherited its implementation.}. 
We implement an architecture like the Feature Pyramid Network (FPN)~\cite{supp_FPN}, illustrated in \fref{fig:supp_two_stage_arch}, to decode the node \& edge maps from the image features.
\Fref{fig:PAF_conf} visually compares the estimated node \& edge maps, showing a better accuracy by our two-stage implementation.

The original ViNet estimates multiple classes of nodes (four classes) and edges (five classes) as different maps for the grapevine dataset. Compared to just using binary classes (\ie, a branch exists or not), our two-stage implementation also yields better estimation accuracies using the multiple classes (SMD in $4.2 \times 10^{-4}$ with multi-class and $1.4 \times 10^{-2}$ with binary classes). Therefore, we use the multi-class setup for our two-stage implementation of the grapevine dataset. For the other dataset, we use binary classification since we do not have specific class information.

\paragraph{Second stage: Graph algorithm}
Given the node confidence and edge direction maps, ViNet~\cite{Guyot} first extracts the node positions, followed by the computation of the \emph{resistivity} between each node pair, defined using the edge directions and the Euclidean distance between nodes. The final estimates of the graph structure are generated using the Dijkstra algorithm, where the tree structure is obtained by computing the shortest paths from all nodes to the detected root crown. The resistivity is used as the edge cost for the Dijkstra algorithm.

For the second stage, we follow the method in \cite{Guyot} except for the graph algorithm used; namely, we compute MST instead of the shortest paths given by the Dijkstra algorithm, since using MST reduces the SMD metric to $4.2 \times 10^{-4}$, compared to $5.9 \times 10^{-4}$ using the Dijkstra algorithm for the grapevine dataset.

\paragraph{Detailed parameter settings}
The two-stage method involves heuristic parameters for node and edge detection. Therefore, we empirically select the best parameter sets for each dataset. \Tref{tab:supp_params} lists the detailed parameters. In particular, for the root dataset, we need to carefully tune some hyperparameters (namely, $d$, $\tau_m$, and $\tau_n$ in the table) to yield reasonable estimates, where the configurations yielding the best SMD scores are reported in the main paper. 

\subsection{Test-time constraint baseline}
\label{sec:supp_test_time}
For the test-time constraint baseline, we apply MST only in the inference phase, where the graph generator is trained using the same procedure as the unconstrained method. The MST used in this baseline method is identical to our proposed one. 

\section{Performance Analysis}
\label{sec:performance_analysis}
In this section, we present a detailed analysis of the performance of our proposed method. The effectiveness of our method is evaluated through comprehensive experiments in different scenarios. Specifically, we compare our method with the Auto-regressive (AR) model, and we also analyze the performance when our method is applied solely during the training processes. 

\subsection{Comparison with auto-regressive (AR) method}
\label{sec:auto}
While we implement the tree-graph constraint on the state-of-the-art non-autoregressive graph generator, RelationFormer~\cite{Relationformer}, other choices of constrained graph generation are viable. 
Existing works aiming for tree-constrained graph generation, such as in molecule structure estimation~\cite{SpanningTreeMolecules, TreeMolecules_father, TreeMolecules_1}, use auto-regressive (AR) graph generation.
AR methods are a simpler choice for imposing the constraint since it is relatively straightforward to implement the tree-graph constraint in their graph development process. However, since the AR methods generate graph nodes and edges progressively, they are prone to breakdowns due to changes in the output order or errors during the generation. This tendency is particularly pronounced for relatively large graphs, including our setup.

To assess the potential of AR methods, we test the state-of-the-art transformer-based AR graph generator, Generative Graph Transformer (GGT)~\cite{GGT}. \Tref{tab:GGT} compares our method and several variances of GGT on the synthetic tree pattern dataset. The results show that the accuracy of GGT falls short compared to our method, although the vanilla GGT (the top row) mostly outputs tree graphs ($92~\%$) without explicitly imposing the tree-graph constraint. We identified that errors by GGT occurring at a particular step in the AR generation process continuously cause errors in the sequence of following generations. The GGT was initially designed for small datasets, specifically for graphs with $\left | V \right | \le 10$. For our setup, where $\left | V \right | \geq 100$, generating these long sequences in a specific order presents a significant challenge.

\begin{table}[t]
\centering
\caption{Comparisons of different graph generation models (\ie, RelationFormer~\cite{Relationformer} and GGT~\cite{GGT}) on the synthetic dataset. }
\label{tab:GGT}
\resizebox{\linewidth}{!}{
\begin{tabular}{c|c|ccc|c}
\hline
 \multirow{2}{*}{Method}& \multirow{2}{*}{SMD $\downarrow$} & \multicolumn{3}{c|}{TOPO score $\uparrow$} & Tree rate \\
                                             &                           & Prec.     & Rec.     & F1       & [\%]   \\ \hline
 GGT~\cite{GGT}                & $2.71 \times 10^{-3}$     & 0.635     & 0.537    & 0.582    & 92.06     \\
 GGT w/ test-time constraint   & $4.13 \times 10^{-3}$     & 0.620     & 0.545    & 0.580    & 98.10     \\
 GGT w/ SFS layer              & $2.80 \times 10^{-3}$     & 0.652     & 0.584    & 0.616    & 99.63    \\ \hline 
 RelationFormer~\cite{Relationformer} w/ SFS layer (Ours)& $4.78 \times 10^{-6}$     & 0.986     & 0.968    & 0.977    & 100.0      \\ \hline
\end{tabular}
}
\end{table}

\subsection{Effectiveness of tree-constraint during training}
To assess whether our SFS layer (positively) affects the training process itself or not, we evaluate our method \emph{without} using the SFS layer and the MST algorithm during the inference phase, \ie, introducing constraint only during the training phase (called  \emph{train-time constraint} hereafter). \Tref{tab:traintime} summarizes the performances. Inducting the tree constraint during the training phase mostly outperforms the methods without constraints, meaning that the improvement by our method is based on network improvement by the loss propagated via the SFS layer. 
We also checked the change in the accuracy metric during training, and found our method consistently achieved better accuracy from the beginning of the training. 

\begin{table}[t]
\centering
\caption{Quantitative results with additional baseline, \emph{train-time constraint}.}
\label{tab:traintime}
\resizebox{\linewidth}{!}{
\begin{tabular}{c|c|c|c|ccc|c}
\toprule
\multirow{2}{*}{Dataset}   & \multirow{2}{*}{Method} & \multirow{2}{*}{SFS} & \multirow{2}{*}{SMD $\downarrow$} & \multicolumn{3}{c|}{TOPO score $\uparrow$} & Tree rate \\
                           &                      &                &                           & Prec. & Rec. & F1              & [\%]   \\ \midrule 
\multirow{4}{*}{Synthetic} & Unconstrained        &                & $1.43 \times 10^{-5}$     &0.978         &0.929         &0.953            &36.2      \\ 
                           & Test-time constraint &                & $6.26 \times 10^{-6}$     &0.977         &0.953         &0.965           &\textbf{100.0}     \\
                           & Train-time constraint& \checkmark     & $8.44 \times 10^{-6}$     &\textbf{0.987}&0.954         &0.970           &56.5     \\
                           & Ours                 & \checkmark     & \bm{$4.78 \times 10^{-6}$}&0.986         &\textbf{0.968}&\textbf{0.977}  &\textbf{100.0}     \\ \midrule
\multirow{4}{*}{Root}      & Unconstrained        &                & $1.19 \times 10^{-4}$     &0.831         &0.633         &0.719           &35.9     \\
                           & Test-time constraint &                & $1.52 \times 10^{-4}$     & 0.829         &0.771        &0.799           &\textbf{100.0}     \\
                           & Train-time constraint& \checkmark     & \bm{$7.81 \times 10^{-5}$}&0.853         &0.619         &0.718          &37.2     \\
                           & Ours                 & \checkmark     & $8.82 \times 10^{-5}$     &\textbf{0.861}&\textbf{0.807}&\textbf{0.833}   &\textbf{100.0}  \\ \midrule
\multirow{4}{*}{Grapevine} & Unconstrained        &                & $1.45 \times 10^{-4}$     &0.963         &0.559         &0.708           &0.0      \\
                           & Test-time constraint &                & $1.47 \times 10^{-4}$     &0.896         &0.840         &0.867            &\textbf{100.0}     \\
                           & Train-time constraint& \checkmark     & $1.30 \times 10^{-4}$     &\textbf{0.965}&0.566         &0.713           &0.0     \\
                           & Ours                 & \checkmark     & \bm{$1.03 \times 10^{-4}$}&0.899         &\textbf{0.843}&\textbf{0.870}  &\textbf{100.0}     \\ \bottomrule
\end{tabular}
}
\end{table}

\section{Other Design Choices}
\label{sec:supp_ablation}
The experiments in the main paper already provide some ablation studies, namely, comparisons of our method with 1) graph generation without constraint (\emph{unconstrained}), and 2) a method without using MST in the training loop (\emph{test-time constraint}). Here, we delve further into the potential design choices of our TreeFormer model.

\subsection{Other graph generators}
Although the proposed module, the SFS layer, can be easily integrated into graph generators other than RelationFormer~\cite{Relationformer}, we found that no methods but our TreeFormer implementation achieve satisfactory results. Here, we discuss results by the implementation of our method to the AR graph generator, GGT~\cite{GGT}, which achieves the second-best accuracy for multiple datasets following the state-of-the-art RelationFormer.

\Tref{tab:GGT} in the last section compares the GGT with and without the tree-graph constraint. Compared to the GGT with test-time MST, using our SFS layer on top of GGT improves both SMD and TOPO scores\footnote{GGT w/ SFS layer does not achieve $100$~[\%] tree rate because it sometimes fails to generate any graphs.}, although the accuracies are insufficient in practice due to the drawback of AR-based generation processes discussed above. Using the newer RelationFormer model significantly improves the estimation accuracy, which implies that our SFS layer will benefit from the future development of graph generation models.

\subsection{Using node distances for edge cost in MST}
Although our proposed method uses the edge non-existence probabilities $\{\unconst{y}^{-}_{\ijsub}\}$ as the edge cost for the MST algorithm, inspired by the two-stage method that uses node distance for the edge cost computation, we multiply the Euclidean distance between nodes by our original edge cost.

As a result, SMD with modified edge cost does not improve accuracy (it achieves the same SMD as our method in the Grapevine dataset).
A possible reason for this is that the graph generator itself can take the node distance into account when estimating graph edges.
Therefore, we simply use the edge non-existence probabilities $\{\unconst{y}^{-}_{\ijsub}\}$ as the edge cost for our method. 

\begin{table}
\centering
\caption{Ablation for $\Lambda$.}
\label{tab:ablation}
\resizebox{\linewidth}{!}{
\begin{tabular}{c|c|ccc}
\hline
 \multirow{2}{*}{$\Lambda$}& \multirow{2}{*}{SMD $\downarrow$} & \multicolumn{3}{c}{TOPO score $\uparrow$}  \\
                              &                           & Prec.     & Rec.     & F1         \\ \hline
$2$ $(\exp(-\Lambda) = 1.4\times 10^{-1})$         & $1.51 \times 10^{-4}$     & $0.871$     & $0.803$    & $0.836$         \\ 
$5$ $(\exp(-\Lambda) = 6.7\times 10^{-3})$         & $1.27 \times 10^{-4}$     & $0.866$     & $0.799$    & $0.831$         \\ 
$10$ $(\exp(-\Lambda) = 4.5\times 10^{-5})$         & $\mathbf{1.03 \times 10^{-4}}$     & $\mathbf{0.899}$     & $\mathbf{0.843}$    & $\mathbf{0.870}$         \\ 
$100$ $(\exp(-\Lambda) = 3.7\times 10^{-44})$       & $1.07 \times 10^{-4}$     & $0.886$    & $0.830$    & $0.857$        \\ \hline
\end{tabular}
}
\end{table}

\subsection{Ablation for $\Lambda$}
An important hyperparameter in our method is $\Lambda$, which controls the level of suppression for unwanted features.
Here, we report an ablation study for this parameter using the grapevine dataset. 
\Tref{tab:ablation} shows that our choice ($\Lambda=10$) achieves better, while the changes in $\Lambda$ do not significantly affect the overall accuracy as long as $\exp(-\Lambda)$ is close enough to zero. This result indicates that our method is robust to the hyperparameter setting.

\section{Additional Visual Results}
\label{sec:supp_results}
\begin{figure*}[tp]
	\centering
  \includegraphics[width=\textwidth,height=\dimexpr\textheight-2\baselineskip\relax,keepaspectratio]{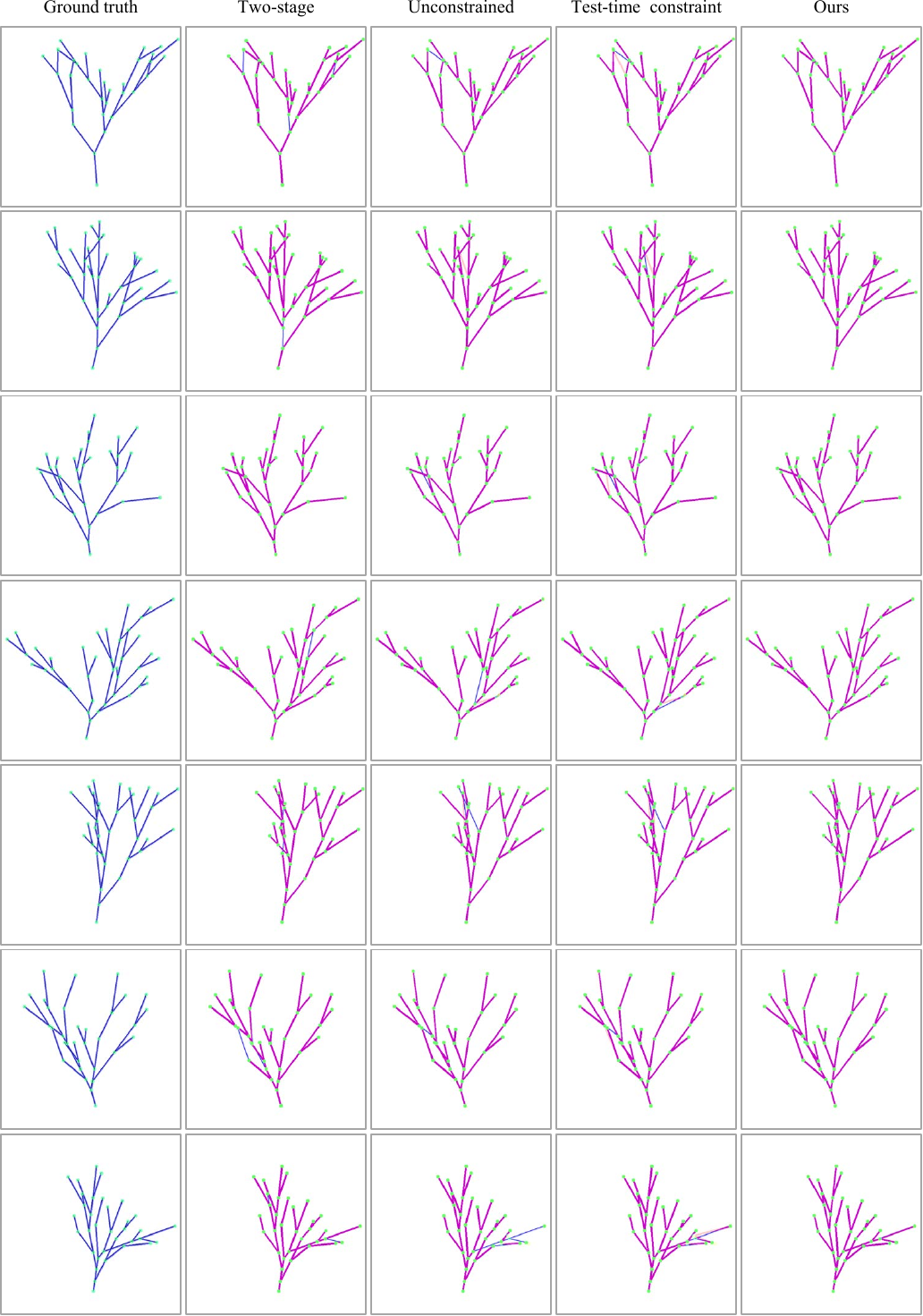}    
	\caption{Additional results for the synthetic branch pattern dataset.}
    \label{fig:supp_result_syn}
\end{figure*}

\begin{figure*}[tp]
	\centering
  \includegraphics[width=\textwidth,height=\dimexpr\textheight-2\baselineskip\relax,keepaspectratio]{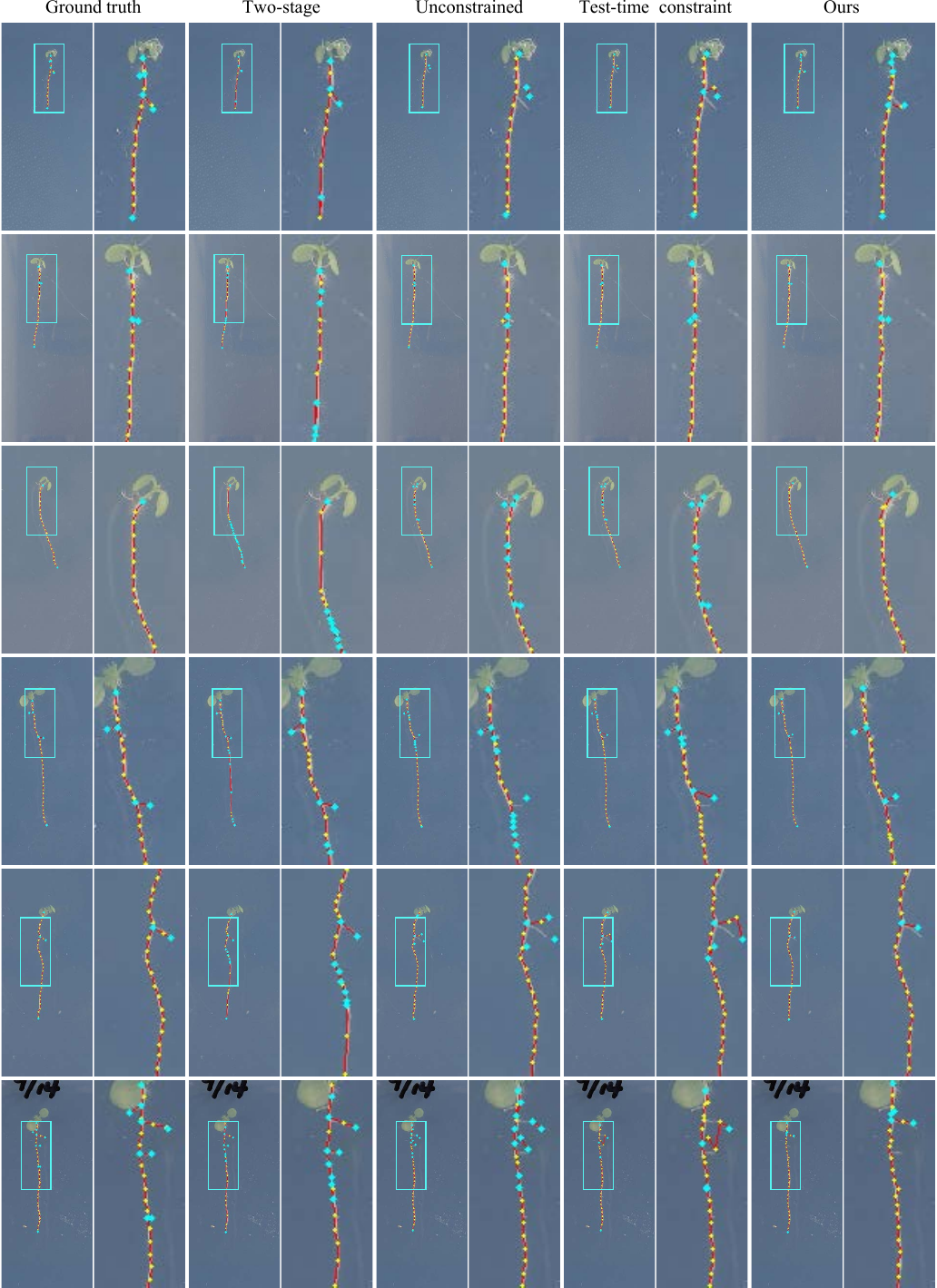}    
	\caption{Additional results for the root dataset.}
    \label{fig:supp_result_root}
\end{figure*}

\begin{figure*}[tp]
	\centering
  \includegraphics[width=\textwidth,height=\dimexpr\textheight-2\baselineskip\relax,keepaspectratio]{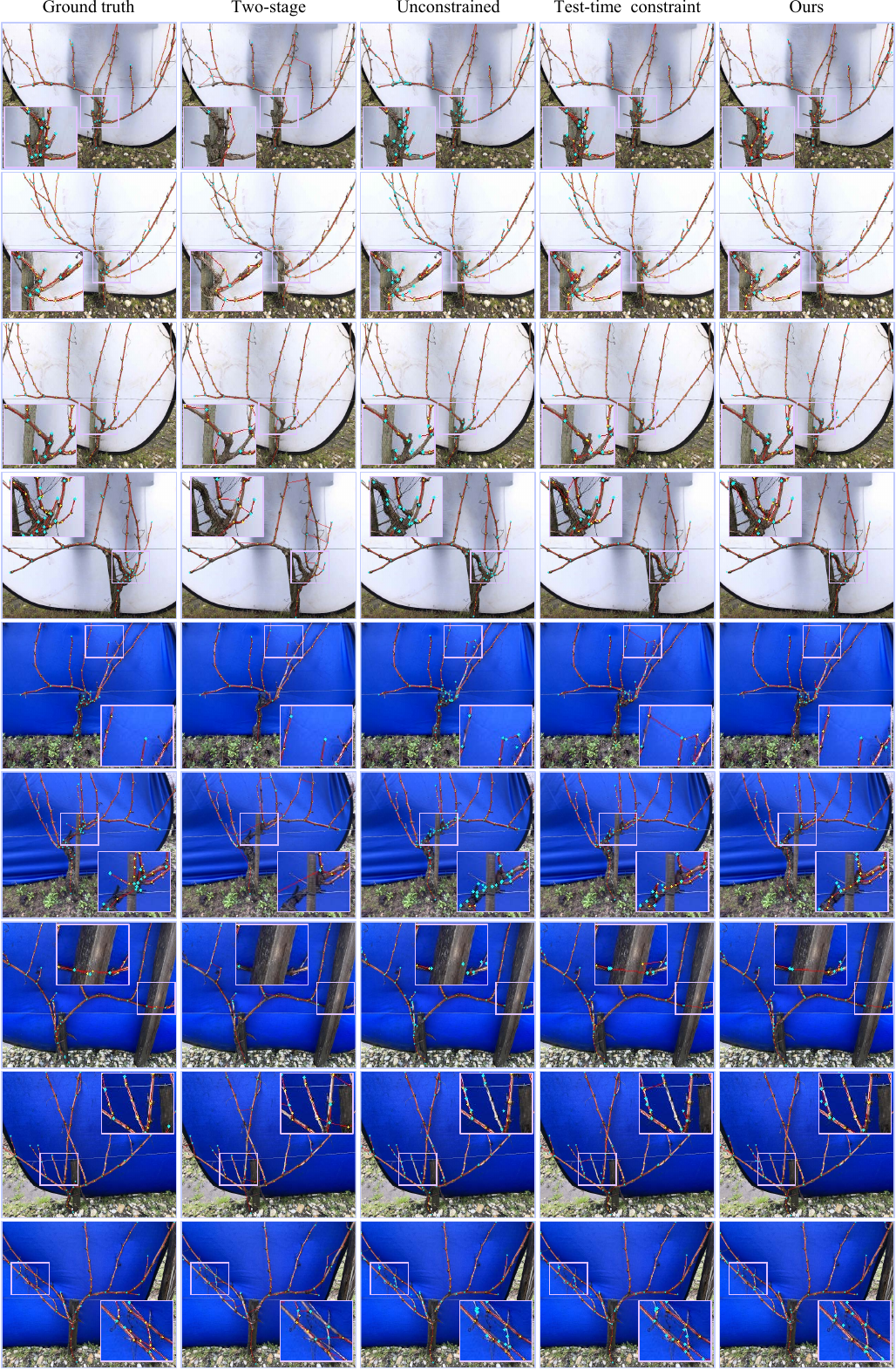}    
	\caption{Additional results for the grapevine dataset.}
    \label{fig:supp_result_grapevine1}
\end{figure*}

\begin{figure*}[tp]
	\centering
  \includegraphics[width=\textwidth,height=\dimexpr\textheight-2\baselineskip\relax,keepaspectratio]{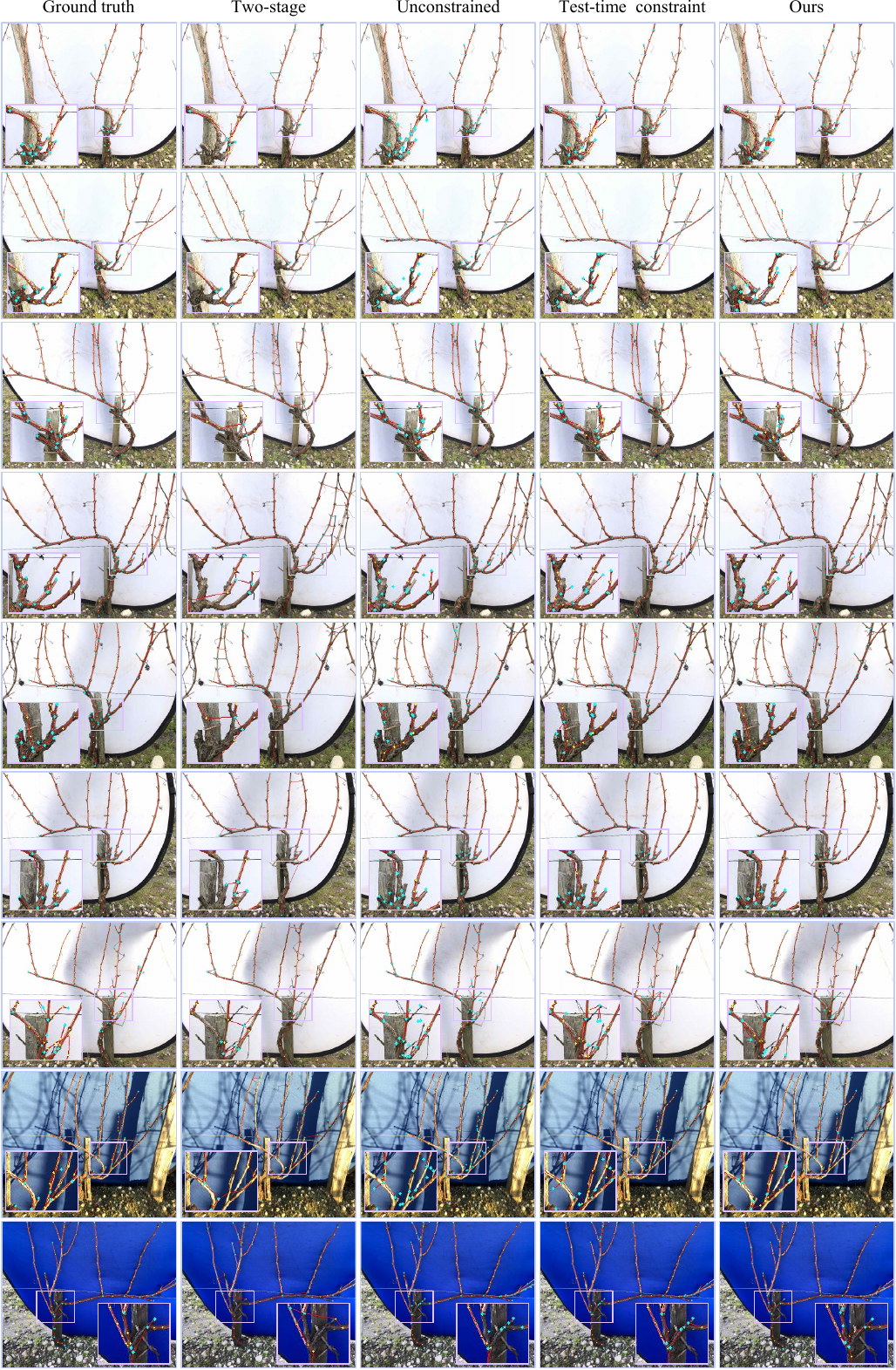}    
	\caption{Additional results for the grapevine dataset (cont'd).}
    \label{fig:supp_result_grapevine2}
\end{figure*}

\begin{figure*}[t]
	\centering
  \includegraphics[width=\textwidth,height=\textheight,keepaspectratio]{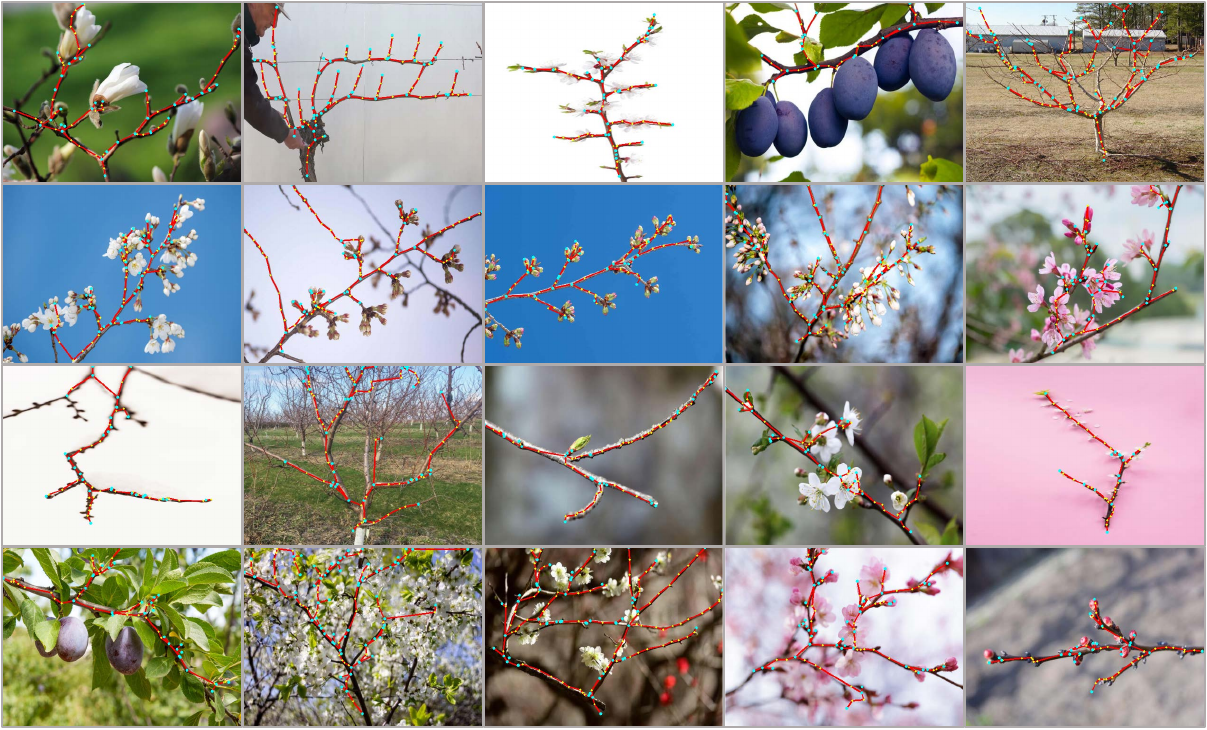}    
	\caption{Additional results for the out-of-domain test dataset.}
    \label{fig:supp_result_domain}
\end{figure*}

We finally show additional visual results. 
Figures~\ref{fig:supp_result_syn} and \ref{fig:supp_result_root} show the additional results for synthetic and root datasets, respectively.
Figures~\ref{fig:supp_result_grapevine1} and \ref{fig:supp_result_grapevine2} show the results for the grapevine dataset.
Figure~\ref{fig:supp_result_domain} show the results for out-of-domain testing.

These results consistently demonstrate the high-fidelity estimation of plant skeletons by our TreeFormer, which uses the SFS layer that incorporates the constraints while training graph generation models.

{\small
\bibliographystyle{ieee_fullname}
\bibliography{egbib}
}

\end{document}

%% file: mlpreamble.tex
\newcommand{\Tref}[1]{Table~\ref{#1}}
\newcommand{\eref}[1]{Eq.~\eqref{#1}}

\newcommand{\fref}[1]{Fig.~\ref{#1}}
\newcommand{\Fref}[1]{Figure~\ref{#1}}
\newcommand{\sref}[1]{Sec.~\ref{#1}}

\usepackage{xcolor}
\newcounter{todos}
\AtEndDocument{\ifnum\value{todos}>0 \PackageWarning{TODOS}{There are \arabic{todos} todos left in this paper! Fix them before submitting the paper!} \fi}

\newcommand{\V}[1]{\ensuremath{\mathbf{#1}}}

\usepackage{bm}

